# Systematic review of deep learning and machine learning for building energy

**Ardabili Sina[1], Leila Abdolalizadeh[1], Csaba Mako[2], Bernat Torok[2], Mosavi Amir[3,4,5]***

[1] Department of Informatics, J. Selye University, Komarom, Slovakia

[2] Institute of Information Society, University of Public Service, Budapest, Hungary

[3] Faculty of Civil Engineering, Technische Universitat Dresden, Dresden, 01069, Germany

[4] Institute of Information Engineering, Automation and Mathematics, Slovak University of Technology in Bratislava, Slovakia

[5] John von Neumann Faculty of Informatics, Obuda University, Budapest, Hungary

* Correspondence: Amir Mosavi (amir.mosavi@kvk.uni-obuda.hu)

## Abstract

The building energy (BE) management has an essential role in urban sustainability and smart cities. Recently, the novel data science and data-driven technologies have shown significant progress in analyzing the energy consumption and energy demand data sets for a smarter energy management. The machine learning (ML) and deep learning (DL) methods and applications, in particular, have been promising for the advancement of the accurate and high-performance energy models. The present study provides a comprehensive review of ML and DL-based techniques applied for handling BE systems, and it further evaluates the performance of these techniques. Through a systematic review and a comprehensive taxonomy, the advances of ML and DL-based techniques are carefully investigated, and the promising models are introduced. According to the results obtained for energy demand forecasting, the hybrid and ensemble methods are located in high robustness range, SVM-based methods are located in good robustness limitation, ANN-based methods are located in medium robustness limitation and linear regression models are located in low robustness limitations. On the other hand, for energy consumption forecasting, DL-based, hybrid, and ensemble-based models provided the highest robustness score. ANN, SVM, and single ML models provided good and medium robustness and LR-based models provided the lower robustness score. In addition, for energy load forecasting, LR-based models provided the lower robustness score. The hybrid and ensemble-based models provided a higher robustness score. The DL-based and SVM-based techniques provided a good robustness score and ANN-based techniques provided a medium robustness score.

## 1 Introduction

One of the essential aspect of smart buildings is to provide optimum living conditions following numerous standards and energy parameters (Al Dakheel and Tabet Aoul 2017, Li, Ding et al. 2017). There is a need for energy consumption management to achieve optimum comfort, cost and tranquility in the buildings (Vázquez-Canteli, Ulyanin et al. 2019). Therefore, several smart facilities and equipment are installed in the building to regulate the living condition. The major parts of energy consumption in buildings provide thermal and refrigeration comfort, i.e., cooling and heating systems, water supply facilities, sanitary spas, lighting-related facilities, and so on (Ardabili, Mahmoudi et al. 2016, Ma, Saha et al. 2017). In addition, different equipment is installed in each building depending on the type of building, each of which in turn consumes energy. Therefore, in each building, energy is used in different ways to face with the needs of residents. Buildings are candidate for about forty percent of the total energy consumption (Vázquez-Canteli, Ulyanin et al. 2019). The management of the energy in buildings can be considered as one of the important aspects of the smart cities. The sustainability index in urban development can be carefully considered as a social function of the energy generation and local consumption in each developing city. The sustainability index of urban development efficiently is a function of energy generation and direct consumption of each developing city. The energy consumption of buildings is responsible for a considerable value of energy consumption in an urban settlement (Ardabili, Mosavi et al. 2019). There are several techniques aiming to accurately estimate and predict the energy production and consumption in the building sector (Mocanu, Nguyen et al. 2016, Li, Ding et al. 2017, Singaravel, Suykens et al. 2018). In overall, at the building level, two types of necessary actions in this direction can be helpful. A series of models are based on physical principles to justify thermal dynamics and energy behavior in mathematical language. These models are the basic models and characterized based on the type of building and effective parameters. Models developed based on statistical methods are other types of models that are used to estimate the energy consumption based on variables affecting climate and energy costs. From this perspective the forecasting of demand and consumption is important in development of smart cities. Machine learning (ML) based techniques as a subset of Artificial Intelligence (AI), can provide a practical platform for modeling by considering a wide range of parameters (Fan, Xiao et al. 2017). ML based techniques have recently contributed significantly in implementing the reliable estimation models (Amasyali, El-Gohary et al. 2018, Zou, Zhou et al. 2018, Cai, Pipattanasomporn et al. 2019, Zhang, Tian et al. 2020). Several researchers have employed ML-based techniques in different fields of area (Safdari Shadloo, Rahmat et al. 2021). Yahya et al. (2021) employed multi-layered-perceptron (MLP), radial basis function (RBF), cascade feedforward (CFF), and generalized regression neural networks (GRNN) for estimation of the thermal conductivity of water-alumina nano-fluids (Yahya, Rezaei et al. 2021). Zhang et al. (2021) employed an AI-based optimization procedure to obtain the highest efficiency of an off-grid hybrid renewable energy scheme composed of wind, fuel cell, and hydrogen storage schemes (Zhang, Maleki et al. 2021). Therefore, these techniques cover a wide range of scientific fields. Fig. 1 claims the exponential trend in implementing the ML based techniques in this realm during the past decade. The main contribution of the present study is to peruse the use of novel ML based techniques in forming the applications of smart cities in terms of energy. The scope covers building energy (BE) demand and consumption, and energy load prediction which are the key energy concerns in the building sector (Chou and Ngo 2016, Fan, Wang et al. 2019, Zhang, Li et al. 2020).

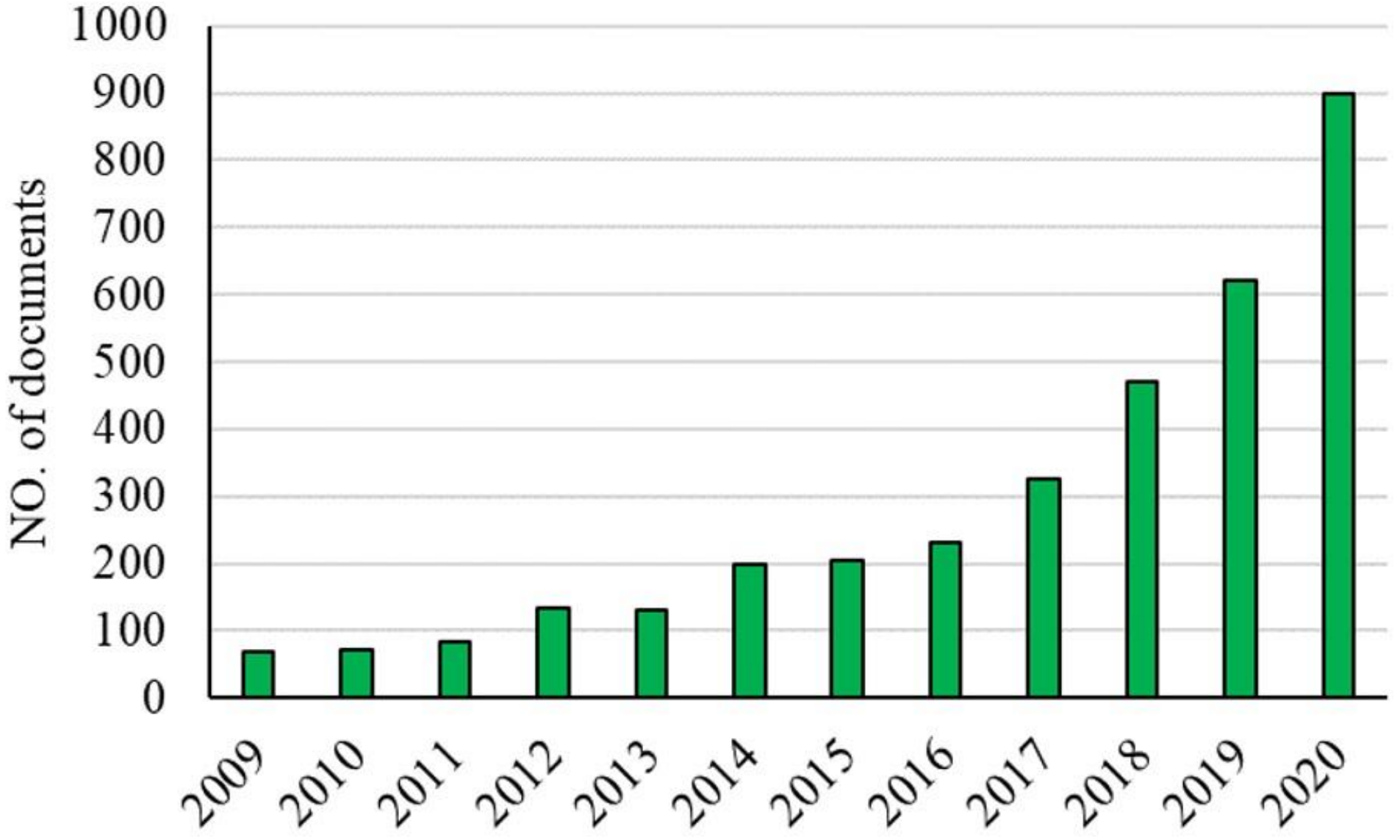

**Figure 1.** Number of documents

In this way there are several survey papers which have been developed with a little overlap in terms of the purpose of the study. Mrtinez et al (2021) developed a survey paper for analyzing the AI-based approached employed in distribution networks (Barja-Martinez, Aragüés-Peñalba et al. 2021). Hasan and Roy (2021) developed a review work by emphasizing on the application of DL-based techniques, transfer learning, active learning, and reinforcement learning for handling the cyber-physical building environment energy consumption (Hasan, Roy et al. 2021). In a review work by Mohapatra et al (2021), ML-based and DL-based techniques are evaluated only for predicting the BE consumption (Mohapatra, Mishra et al. 2021). Uzum et al (2021) in a review work evaluated the possible solutions with AI, DL and ML-based optimization techniques for analyzing the effects of rooftop PV on distribution network (Uzum, Onen et al. 2021). Vázquez-Canteli et al (2019) developed a survey study to analyze the reinforcement learning techniques applied for demand response applications in the smart grid (Vázquez-Canteli, Ulyanin et al. 2019). Panchalingam and Chan (2019) developed a survey study for describing and analyzing the energy use in smart buildings using the applications of ML and DL-based techniques (Panchalingam and Chan 2019). A number of recent surveys studied the application of machine learning in building energy. However, there is still a gap in the presence of a standard review article where PRISMA statement is correctly adapted. In addition, an in-depth investigation on the accuracy of the model accuracy is still missing. Other difference is the taxonomy content, study concept and the perspective of the study as well as the main findings of the study. The limitations of the concepts used, as well as the study of a limited community of applications of ML-based techniques in previous studies, led to a more comprehensive study in the present survey to cover the weaknesses of previous studies. One of the main weaknesses in previous studies was the evaluation of ML-based and DL-based techniques in energy applications in building sectors. On the other hand, according to studies, the existence of a systematic review article based on a standard method that can extract all the strengths and weaknesses of using machine learning methods and deep learning in energy applications in the building, is missing. PRISMA method is a method of preparing review articles that has not been discussed in this field. Accordingly, the present study presents a comprehensive systematic review based on the standard PRISMA method. The present study has taken steps to cover these weaknesses by providing evaluation methods and comparing different models in different applications. Accordingly, the present work has four main steps. The first step is to describe the main procedure of the searching and methodology for developing the base of the study. The second step is to analyze the studies with different ML and DL-based techniques separately in different applications. The third step is to analyze the results and main findings including advantages and dis-advantages of each method in each application and the final step is to summarize and conclude the main findings and suggestions of the study.

## 2 Methodology

This section qualifies how to study and the original taxonomy of the work. Given that this study is a review study, it is necessary to collect similar and related studies and categorize them according to the taxonomy of the work. Therefore, before explaining how to collect information and similar studies, we need to explain the main taxonomy of the article. This article is structured in eight main parts. The first part is the introduction so that he can have an initial introduction of the work and present the problem statement, the existing gaps in this field, and the purpose and justify the work. The second part is the methodology section, which explains how to do the study. The main mission of this article begins with the third section. The third section describes the studies conducted in the field of forecasting energy demand in buildings. This section has six sub-sections. First, the structure and description of artificial neural network (ANN) models, support vector machines (SVM), hybrid models, and ensemble techniques are examined. In each subset, studies in this field are discussed in parallel. The fifth subdivision is related to the introduction of evaluation parameters, and the sixth subdivision is related to the presentation of the results of studies conducted in the field of forecasting energy demand in buildings. It should be noted that the description of the structure of the models is done once and in the next sections, only the studies are reviewed. The fourth section presents studies on energy consumption forecasting in buildings. This section has five subsets to express the studies performed with ANN models, SVM, hybrid models, and ensemble models and the results obtained. Therefore, the results of each section are reviewed and discussed within the same section. The fifth section presents studies on energy load forecasting in buildings. This section has five subsets to express the studies performed with ANN models, SVM, hybrid models, and ensemble models and the results obtained. The sixth section examines studies conducted by deep learning (DL) models in the BE sector. Due to the high importance of DL models, these models were separated from other machine learning models to be examined separately. This section also includes four subsets to express and describes the structure of recurrent neural network (RNN), Long-short term memory (LSTM) and convolutional neural network (CNN) and the results obtained by these studies. The seventh section expresses the discussion and the eighth section expresses the conclusion.

The procedure of data collection for review process adopts the PRISMA standard. According to (Mosavi, Faghan et al. 2020), PRISMA method defines four main levels including: (1) identification, (2) screening, (3) eligibility, (4) inclusion for developing a systematic review. According to the identification stage, an initial search was performed among the databases. Using Thomson Reuters Web-of-Science (WoS) and Elsevier Scopus, 350 of the most relevant articles are identified. The next level is to screening the duplicate articles and choosing the relevant articles according to the title and abstract section. In this level, 70 articles have been identified. The next step is eligibility, to study the full text of articles by authors and to select the relevant articles by considering eligibility for the final review process. In this level 52 articles have been selected for the required evaluations. The final level of the PRISMA technique is to build the database of the study for qualitative and quantitative comparisons. The database of the present study includes 52 articles, for performing the required analyses. Figure 2 presents the flow-chart of forming the database of the current study using the PRISMA technique.

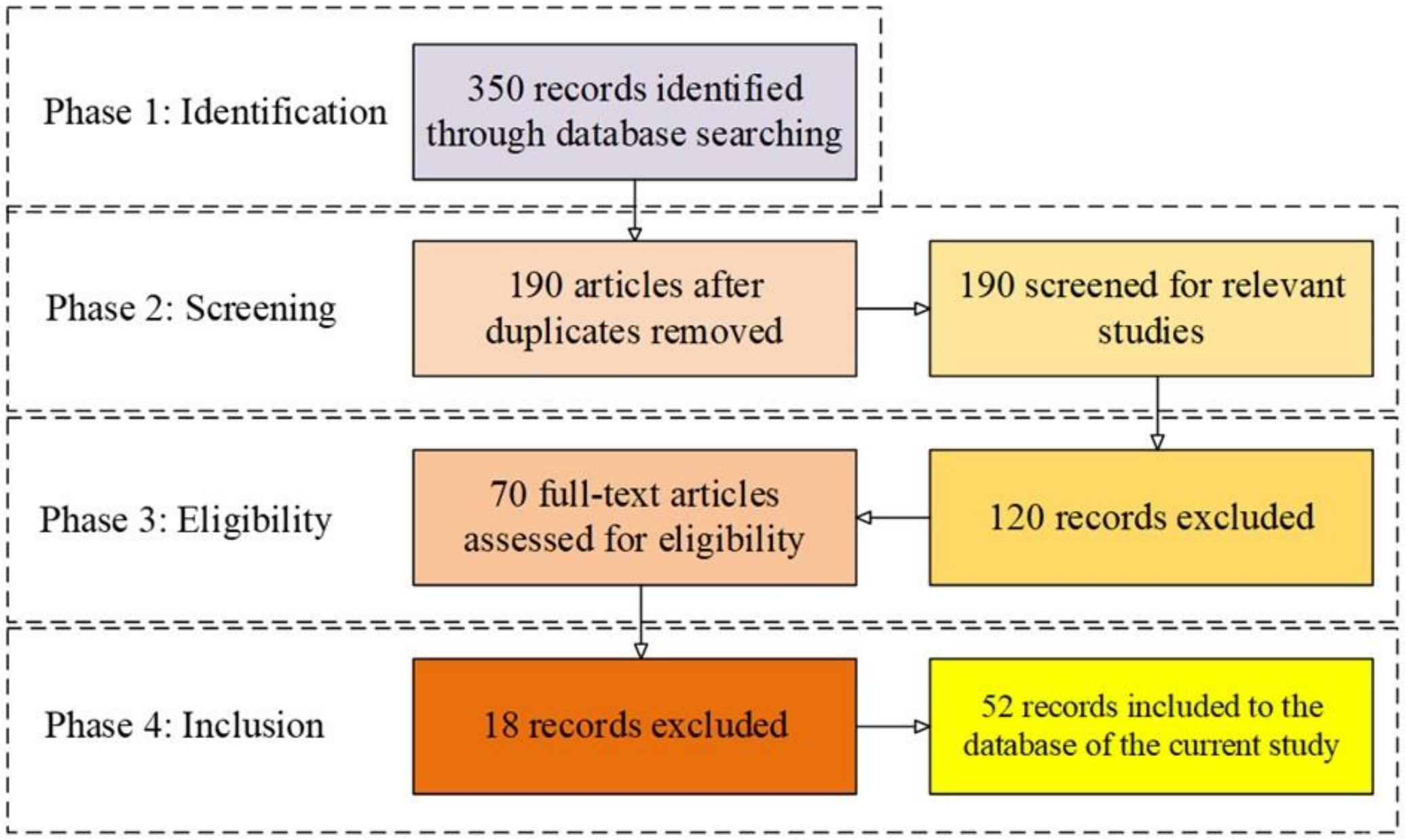

**Figure 2.** The procedure of PRISMA technique for the reviewing process

## 3 Demand prediction in the BE sector

Demand prediction in BE sector is vital for planning and management purposes in the energy systems. This section presents studies developed for demand prediction using different ML methods in the building sector. Table 1 proposes the top studies in the BE demand prediction using ML-based methods.

**Table 1.** Studies developed for BE demand prediction

| Reference | Contribution | Application | ML method(s) | Time period |
|---|---|---|---|---|
| (Ayoub, Musharavati et al. 2018) | To present an ANN predictive model for forecasting micro- | Forecasting | ANN | Short term |
| | level energy supply and demand for buildings | | | |
| (Marmaras, Javed et al. 2017) | To present an ANN model for the estimation of the power demand of the building | Forecasting | ANN | Long term |
| (Buratti, Lascaro et al. 2014) | To present ANN predictive model for the reduction of cooling energy demand in buildings | Optimization | ANN | Short term |
| (Es, Kalender et al. 2014) | To present a forecasting of Turkey energy demand using ANN method | Forecasting | ANN | Long term |
| (Djenouri, Laidi et al. 2019) | A survey for the application of ANN based methods for considering demand problems in smart buildings | Forecasting | ANN, SVM, GA and SVR | Short term |
| (Ahmad, Chen et al. 2018) | A comprehensive review on the application of ANN based techniques for the estimation of energy demand in building sector | Forecasting | ANN and SVM | Short term and Long term |
| (Attanasio, Piscitelli et al. 2019) | SVM method for the prediction of the building Primary Energy Demand in comparison with ANN and DT methods | Forecasting | ANN, SVM and DT | Short term |
| (Paudel, Nguyen et al. 2015) | SVM method for forecasting BE demand in the presence of | Forecasting | SVM | Short term |
| | Pseudo Dynamic Approach | | | |
| (Luo, Oyedele et al. 2019) | To propose a hybrid ML method for day-ahead estimation of BE demands based on IoT-based big data platform | Forecasting | k.means-ANN | Short term |
| (Martina and Amudha 2019) | To develop a hybrid prediction model for the estimation of daily global solar radiation for cope with BE demand | Forecasting | ANFIS | Long term |
| (Kokkinos, Papageorgiou et al. 2017) | To develop hybrid ML techniques for forecasting the energy demand in building sector | Forecasting | ANFIS and ANN | Long term |
| (Popoola, Munda et al. 2015) | To develop a hybrid ML method for the estimation of energy savings associated with energy efficient projects in building sector | Forecasting | ANFIS and ANN | Short term |
| (Raza, Nadarajah et al. 2017) | The estimation of load demand using hybrid ensemble method in BE sector | Forecasting | Ensemble method | Short term |
| (Huang, Yuan et al. 2019) | To propose an ensemble technique for the prediction of energy demand in building sector | Forecasting | Ensemble method, ELM and MLR | Long term |

Ayoub et al. (2018) employed ANN for forecasting micro-level energy supply and demand for buildings (Ayoub, Musharavati et al. 2018). Marmaras et al. (2017) developed an ANN-based technique for the estimation of the power demand of the building (Marmaras, Javed et al. 2017). Buratti et al. (2014) presented an ANN-based predictive model for the reduction of cooling energy demand in buildings (Buratti, Lascaro et al. 2014). Avni et al. (2014) presented a forecasting of Turkey energy demand using ANN method(Es, Kalender et al. 2014). Djenouri et al. (2019) presented a survey for the use of ANN based methods for considering demand problems in smart buildings (Djenouri, Laidi et al. 2019). Attanasio et al. (2019) developed SVM method for the prediction of the building Primary Energy Demand in comparison with ANN and DT methods (Attanasio, Piscitelli et al. 2019). Paudel et al. (2015) developed SVM method for forecasting BE demand in the presence of Pseudo Dynamic Approach (Paudel, Nguyen et al. 2015). Luo et al. (2019) developed a hybrid ML-based method for day-ahead prediction of BE demands based on IoT-based big data platform (Luo, Oyedele et al. 2019). Martina et al. (2019) developed a hybrid prediction model for the estimation of daily global solar radiation for cope with BE demand (Martina and Amudha 2019). Konstantinos et al. (2017) developed hybrid ML-based methods for the prediction of energy demand in building sector (Kokkinos, Papageorgiou et al. 2017). Popoola et al. (2015) developed a hybrid ML-based method for the prediction of energy savings associated with energy efficient projects in building sector (Popoola, Munda et al. 2015). Reza et al. (2017) provided a platform for the estimation of load demand using hybrid ensemble method in BE sector (Raza, Nadarajah et al. 2017). Huang et al. (2019) employed a novel ensemble technique for the estimation of energy demand in building sector (Huang, Yuan et al. 2019). Consequently, the major share and allocation of the studies conducted for the evaluation of the ML-based techniques for the BE sector's applications, are related for forecasting and optimization purposes. results claimed that, the ML-based methods could successfully cope with the task.

### 3.1 ANN based studies

ANN can be considered as the practical and frequently used technique among other computational intelligence techniques. ANN based methods can be successfully applied in forecasting, classification, modeling, clustering, error filtering, and optimization purposes. Artificial neurons make the center unit of ANNs. Components of ANN contain neurons, connections, weights and biases and propagation function. Fig. 3 presents a simple form of an ANN method in the presence of the related compounds. In general, there are three steps for developing an ANN method, training, testing and validating. Training is the most important step because it generates the ANN network.

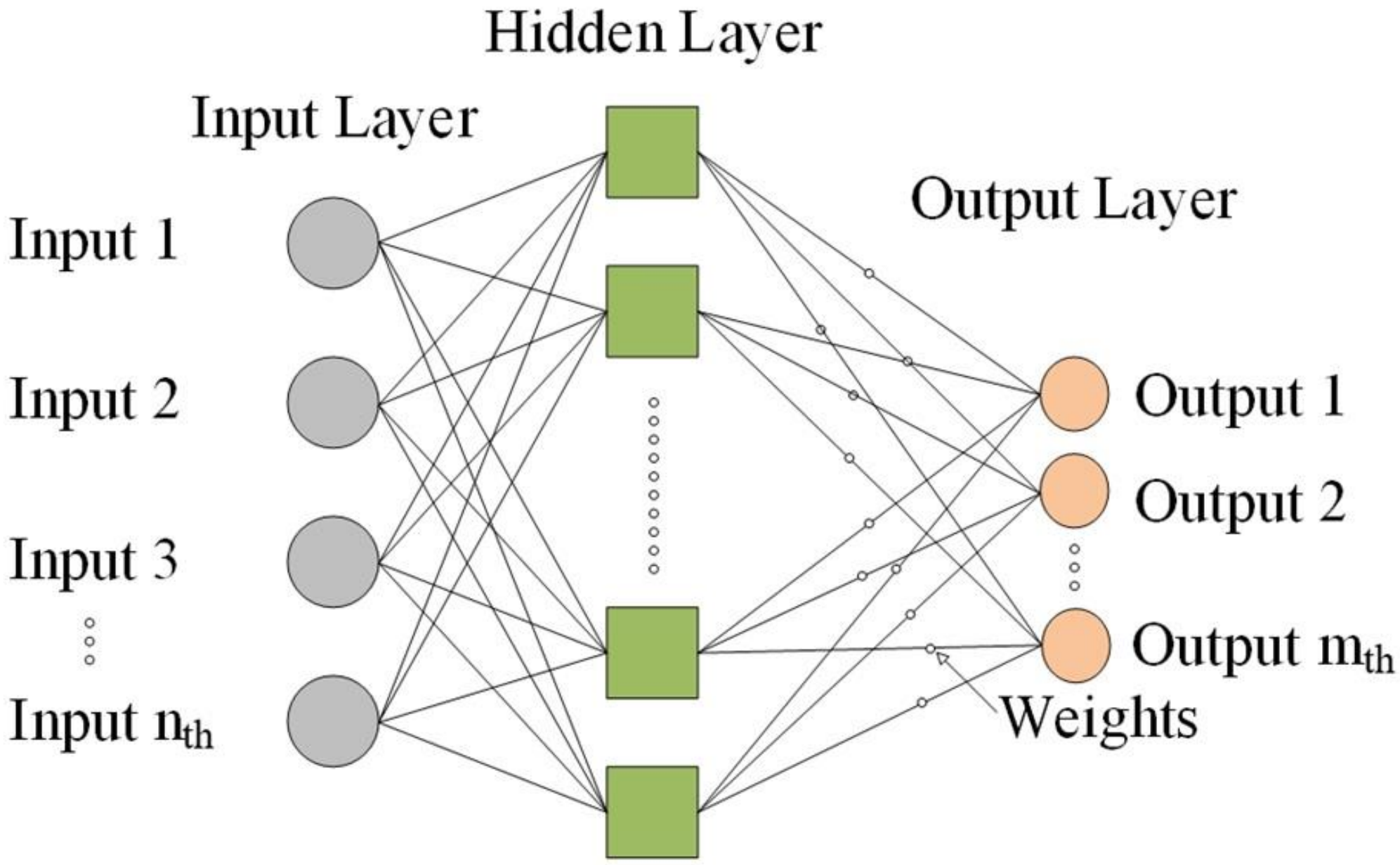

**Figure 3.** The architecture of a simple ANN method

ANN contains connections which transport information from the previous node (which called output of a neuron) to the next node (which called the input of a neuron). Eq. 1 presents the total simple function of the ANN application in the presence of weights and biases.

$$I_j(t) = \sum_i O_i(t) w_{ij} + w_{0j} \quad (1)$$

Where $I_j$ is the input value from the neuron i to neuron j in the presence of $O_i$ as the output value of the neuron i. $w_{ij}$ is the weight value and $w_{0j}$ is the related bias for the neuron j.

Recently, ANN based techniques have been successfully used in BE information sector for controlling, prediction and optimization tasks. Ayoub et al. (Ayoub, Musharavati et al. 2018) developed ANN for the forecasting task in energy demand sector for buildings for considering the building demand in the presence of a hybrid supply system. Accordingly, results claimed that, ANN could cope with the forecasting task. But there was a need for a robust model for preventing overfitting issues. Marmaras et al. (Marmaras, Javed et al. 2017) presented an ANN predictive model for the estimation of power demand in building. Six commercial buildings electricity demand data was employed in order to train the ANN method. Data was collected from a business park during one year. Results claimed that, ANN provided higher accuracy and lower error values in comparing the model outputs and target values. Avni et al. (Es, Kalender et al. 2014) developed ANN method for the estimation of energy demand in Turkey using the population and area of the building. The developed ANN technique has been compared with linear regression in terms of accuracy and the robustness of methods. ANN provided higher accuracy in comparison with that for the LR technique. Buratti et al. (Buratti, Lascaro et al. 2014) performed a study for the optimization of cooling energy demand using ANN method in the presence of indoor thermal comfort. All of these studies have successfully employed ANN to obtain the desired results. However, in further studies, researchers found that ANN based methods can be successful only under certain conditions. So that researchers have compared different methods with ANN to reach an accurate model. The most common of these methods was the support vector based methods.

### 3.2 Support vector based studies

Support-vector based machine learning methods are considered as supervised learning models which analyze data required for classification and regression purposes. Support-vector based machine learning builds a model to assign new examples for training step in the presence of one or more categories. This makes support vector based methods to be a non-probabilistic binary linear classifier. In the following it can be found the mathematical model of a linear algorithm of the support vector based machine learning. By considering a training dataset to be $(\overrightarrow{x_1}), \ldots, (\overrightarrow{x_n})$ . Each $\overrightarrow{x_i}$ indicates a p-dimensional vector. Target is to propose the maximum-margin hyper-plane. Eq. 2 indicates any hyper-plane as the set of the desired points.

$$\vec{w}.\vec{x} - b = 0 \tag{2}$$

where $\vec{w}$ is the normal vector to the hyperplane. $\frac{b}{\|\vec{w}\|}$ demonstrates the offset value of the hyper-plane from the origin along the normal vector.

Originally there are two types of margins, hard and soft margins. In hard margin, the optimization problem is to minimize the $\|\vec{w}\|$ along $y_i(\vec{w}.\vec{x} - b) \geq 1$ for i=1… n. But in soft margin, the optimization problem is a little different. In soft margin Eq. 3 has to be minimized.

$$\left[\frac{1}{n}\sum_{i=1}^{n} \max\left(0,1 - y_i(\vec{w}.\vec{x}_i - b\right)\right] + \lambda\|\vec{w}\|^2 \tag{3}$$

where λ indicates the trade-off between enhancing the margin size and ensuring that the $\vec{x}_i$ lie on the correct side of the margin.

Fig. 4 presents the maximum margin hyper-plane for a support vector based machine trained with samples which are called the support vectors.

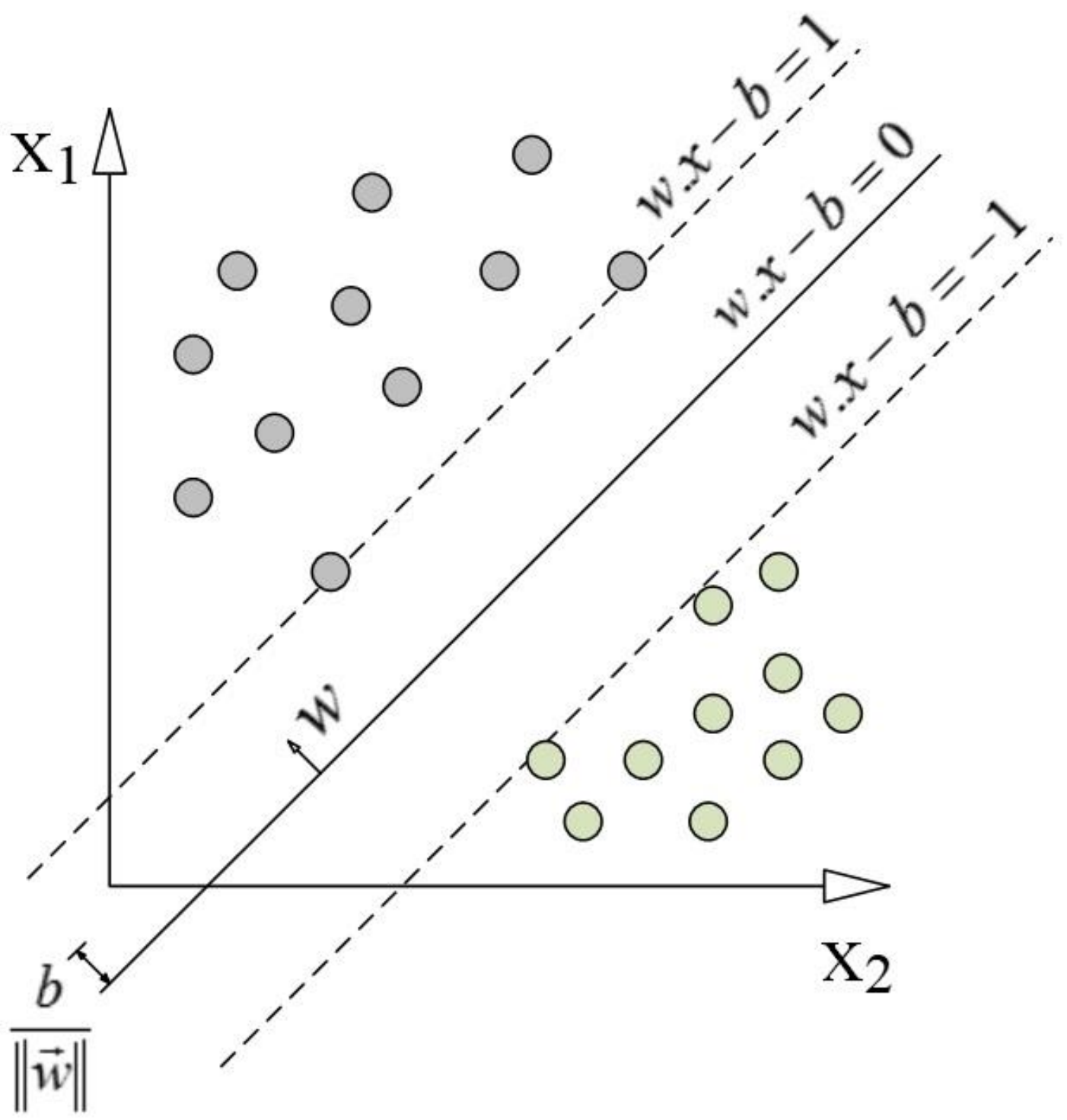

**Figure 4.** The maximum margin hyper-plane for a support vector

Djenouri et al. (Djenouri, Laidi et al. 2019) developed a review work on the use of SVM method in comparison with ANN and GA technique for forecasting energy demand in buildings. SVM method has been proposed for solving issues related to occupants and energy demands. This method has been introduced as a multi-disciplinary solution for energy demand problems in buildings. SVM provided a higher performance compared with ANN but it can be an accurate model in energy demand by developing novel procedures. Ahmad et al. (2018) (Ahmad, Chen et al. 2018) presented a review article in the application of support vector based estimation techniques for forecasting BE demand. Support vector based techniques have been compared with ANN based techniques, as the frequently employed techniques, in terms of accuracy and robustness. The study emphasizes on the highest robustness of support vector based techniques compared with ANN based methods in BE demand sector. Paudel et al. (Paudel, Nguyen et al. 2015) developed a forecasting model for BE demand using SVM. SVM has been introduced as a sensitive model to the selection of training data which made author to employ Dynamic Time Warping for training SVM model.

In another study Attanasio et al. (Attanasio, Piscitelli et al. 2019) developed SVM method for the estimation of the building primary energy demand in comparison with ANN and DT methods for finding an accurate and robust method. The comparison factors were RMSE and accuracy factors. Based on results SVM could successfully do the task with a high reliability and performance compared with ANN method but it performed weak compared with DT method. As a general note, it can be claimed that, in all cases, SVM performs better than ANN method. But, as is reported in the study by Attanasio et al. (Attanasio, Piscitelli et al. 2019), there is another method which can perform better than SVM method for forecasting tasks. RF based methods can be considered with a higher performance compared with ANN and SVM. RF is an ensemble-based learning technique. The following sections presents a brief description about hybrid and ensemble based methods.

### 3.3 Hybrid based studies

Hybrid methods are appeared for generating robust techniques by combining single methods for different purposes such as prediction, classification and optimization purposes. The main aim is to collect the advantages of different methods and eliminate the disadvantage of methods. In the most cases, hybrid methods compose one prediction method as the base method and one optimization method (as complementary or second method) for increasing the accuracy of the prediction method. The most popular hybrid method is ANFIS. ANFIS employs fuzzy rules and ANN architecture for obtaining a sustainable prediction model. Fig. 5 presents a simple algorithm of developing a hybrid method. Recently, these methods have been frequently used in the prediction of energy demand in building sector.

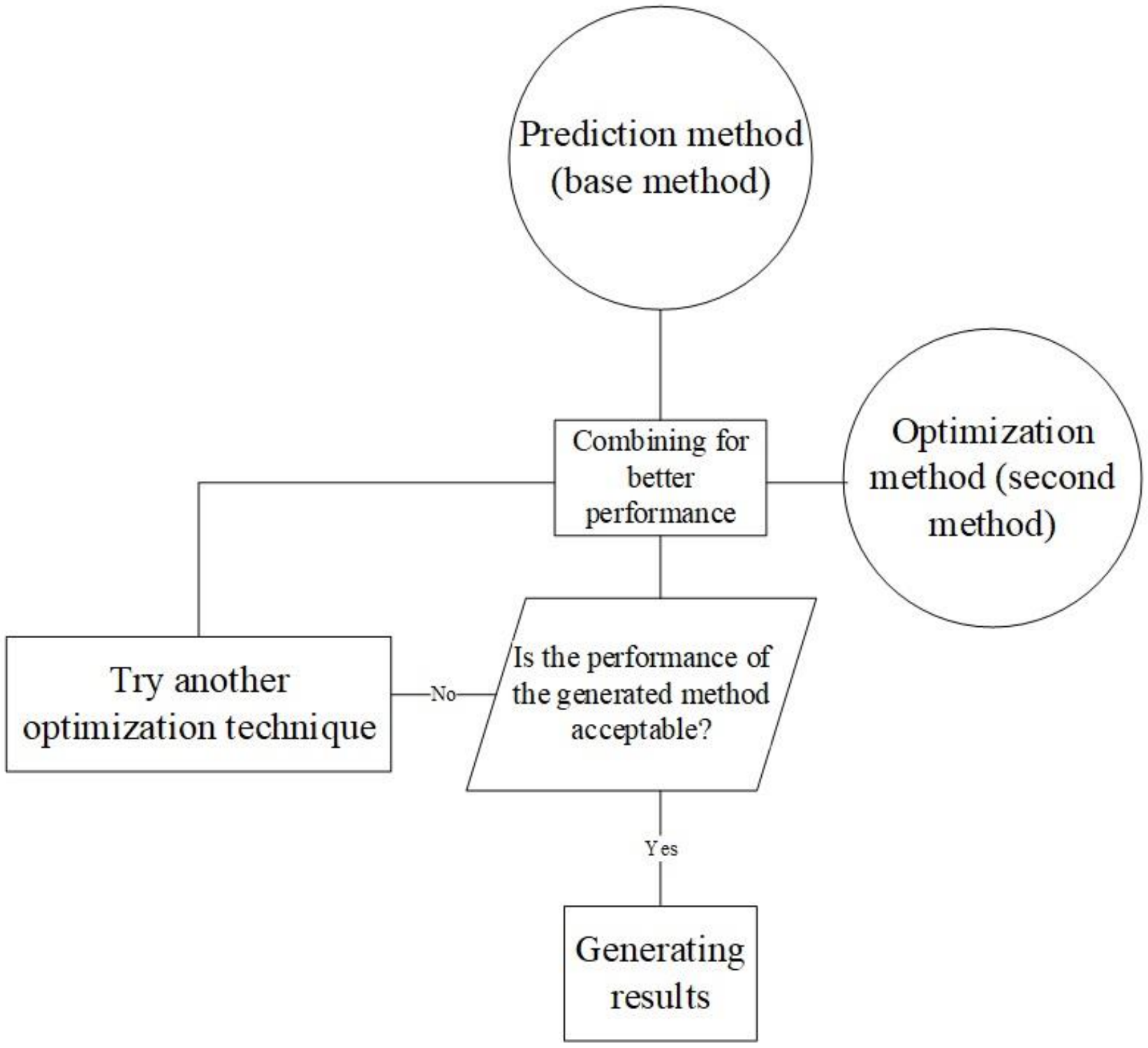

**Figure 5.** A simple algorithm of developing a hybrid technique

Luo et al. (Luo, Oyedele et al. 2019) developed a hybrid k-means-ANN method for the estimation of BE demand using a platform based on IoT-based big data. The proposed method could perform the task in the presence of IoT sensors for training ANN method. Martina et al. (Martina and Amudha 2019) developed an ANFIS method for the prediction of daily global solar radiation in line with the a proper solution for BE demand problems in the presence of meteorological parameters. The proposed ANFIS model has been compared with the statistical models in terms of accuracy and correlation values. In another study, Konstantinos et al. (Kokkinos, Papageorgiou et al. 2017) developed ANFIS method for the prediction of the future energy demand of buildings for obtaining a future perspective form the BE balance. The proposed ANFIS method has been compared with ANN and Fuzzy Cognitive Maps method in terms of RMSE, MAPE and MAE. The performance of ANFIS model was significantly higher than other methods. In another study, Popoola et al. (Popoola, Munda et al. 2015) developed an ANFIS method in comparison with ANN technique for the prediction of energy savings in building sector. The evaluation performance was determination coefficient (R2). ANFIS method provided the best performance due to its hybrid advantages.

### 3.4 Ensemble based studies

Ensemble based methods or in other word multiple classifiers are supervised learning algorithms which have been employed in order to improve the classification accuracies. Developing an efficient ensemble method requires determination of an effective combination of classifiers for elimination of their weakness. These methods have an important difference compared with hybrid methods. Ensemble methods benefits different training algorithms for enhancing the training accuracy for generating a higher performance in the testing phase. Ensemble methods in fact allows for different training algorithms for generating a flexible training (ref. Advances in machine learning modeling, Hybrids and Ensembles). Boosting and bagging are most frequently used ensemble methods. This technique is also frequently employed in energy demand prediction in building sector. Reza et al. (Raza, Nadarajah et al. 2017) developed an ensemble method including neural ensemble, Bayesian model and wavelet transform for the estimation of photo voltaic application in the energy demand for building sector. This technique has been compared in the term of the nRMSE with different single and hybrid machine learning techniques. The selected ensemble method could successfully forecast the demand parameters and enhanced the reliability of the method, significantly. Huang et al. (Huang, Yuan et al. 2019) developed a novel ensemble method by employing extreme gradient boosting, MLR and ELM for developing a SVR method in the presence of historical energy comprehensive variable. Outputs have been evaluated using RMSE and MAE parameters. According to the findings, the proposed ensemble method significantly provided a higher performance compared with single methods.

### 3.5 Criteria for evaluations

The success of the discussed methods has been evaluated based on how capable the developed techniques are in generating most accurate predictions, detection, optimization and monitoring of the process in term of their statistical performance accuracy. Table 2 presents the most common evaluating factors used for comparing the efficiency of the discussed techniques.

**Table 2.** Model evaluation criteria

| Index | Description |
|---|---|
| MSE= $\frac{1}{N}(P-A)^2$ | P=predicted values<br>A=Actual values<br>N=number of data |
| RMSE= $\sqrt{\frac{1}{N}(P-A)^2}$ | |
| MAE= $\frac{1}{N}\lvert P-A\rvert$ | |
| MAPE= $100 \times \frac{1}{N}\lvert\frac{P-A}{A}\rvert$ | |
| Accuracy= $\frac{True\ p+True\ n}{True\ p+True\ n+False\ p+False\ n}$ | p=positive<br>n=negative |
| Reliability= $\frac{\sigma_T^2}{\sigma_X^2}$ | T=True scores<br>X=Errors of measurements |
| Sustainability= $\left\lvert 1-\frac{Testing\ error-Training\ error}{Training\ error}\right\rvert$ | $0 \le Sustainability < 1$ |

MSE = An index to indicate the average of the squares of the deviations from the actual value.

RMSE = An index to indicate the difference between the target and output values.

MAE = An index to indicate the average vertical distance between the target and output values.

MAPE = An index to indicate the relative average vertical distance between the target and output values.

Accuracy = An index for measuring statistical bias value.

Reliability = An index for measuring the overall stability of an experiment

Sustainability= An index for measuring the difference between testing and training errors

## 3.6 Results

Table 3 gives a summarized comparison about the accuracy, reliability and sustainability of techniques developed for the BE demand prediction. Accuracy index has been obtained through the performance indexes related to the training phase and reliability has been obtained from the performance indexes related to the testing phase. But, sustainability index was a little difference and has been obtained by comparing reliability, accuracy, processing time and other factors which have been considered by outputs and findings of the reviewed articles.

**Table 3.** The comparison findings of techniques for BE demand

| **Method** | **Application** | **Accuracy** | **Reliability** | **Sustainability** | **Reference** |
|---|---|---|---|---|---|
| ANN | Regression | ++ | + | + | (Ayoub, Musharavati et al. 2018) |
| ANN | Regression | + | + | + | (Marmaras, Javed et al. 2017) |
| ANN | Optimization | ++ | ++ | + | (Buratti, Lascaro et al. 2014) |
| ANN | Regression | ++ | + | + | (Es, Kalender et al. 2014) |
| ANN | Regression | + | + | + | (Djenouri, Laidi et al. 2019) |
| SVM | Regression | ++ | ++ | ++ | (Djenouri, Laidi et al. 2019) |
| GA | Regression | ++ | + | + | (Djenouri, Laidi et al. 2019) |
| ANN | Regression | + | + | + | (Ahmad, Chen et al. 2018) |
| SVM | Simulation | ++ | ++ | ++ | (Ahmad, Chen et al. 2018) |
| ANN | Simulation | ++ | + | + | (Attanasio, Piscitelli et al. 2019) |
| SVM | Regression | ++ | ++ | ++ | (Attanasio, Piscitelli et al. 2019) |
| DT | Regression | +++ | ++ | ++ | (Attanasio, Piscitelli et al. 2019) |
| SVM | Regression | ++ | ++ | ++ | (Paudel, Nguyen et al. 2015) |
| k.means-ANN | Regression | +++ | ++ | ++ | (Luo, Oyedele et al. 2019) |
| ANFIS | Regression | +++ | ++ | ++ | (Martina and Amudha 2019) |
| ANFIS | Regression | +++ | +++ | ++ | (Kokkinos, Papageorgiou et al. 2017) |
| ANN | Regression | + | + | + | (Kokkinos, Papageorgiou et al. 2017) |
| ANFIS | Regression | +++ | +++ | +++ | (Popoola, Munda et al. 2015) |
| ANN | Regression | ++ | + | + | (Popoola, Munda et al. 2015) |
| Ensemble | Regression | +++ | +++ | +++ | (Raza, Nadarajah et al. 2017) |
| Ensemble | Regression | +++ | +++ | ++ | (Huang, Yuan et al. 2019) |
| ELM | Regression | +++ | ++ | ++ | (Huang, Yuan et al. 2019) |
| MLR | Regression | ++ | + | + | (Huang, Yuan et al. 2019) |

As is clear from table 3, ANN based methods provides the lowest sustainability and hybrid and ensemble based methods provide the highest sustainability. In order to better understand and discuss the power of each method it has been employed an index called robustness. This index has been provided as a novel index for describing the strength of each method based on their accuracy, reliability and sustainability values.

Fig. 6 presents a concluded graph for each technique according to their robustness. Fig. 6 have been categorized into four limitations including high, good, medium and low robustness score to describe the capability and strength of each method based on our own observations and understandings from conclusion and results of each study.

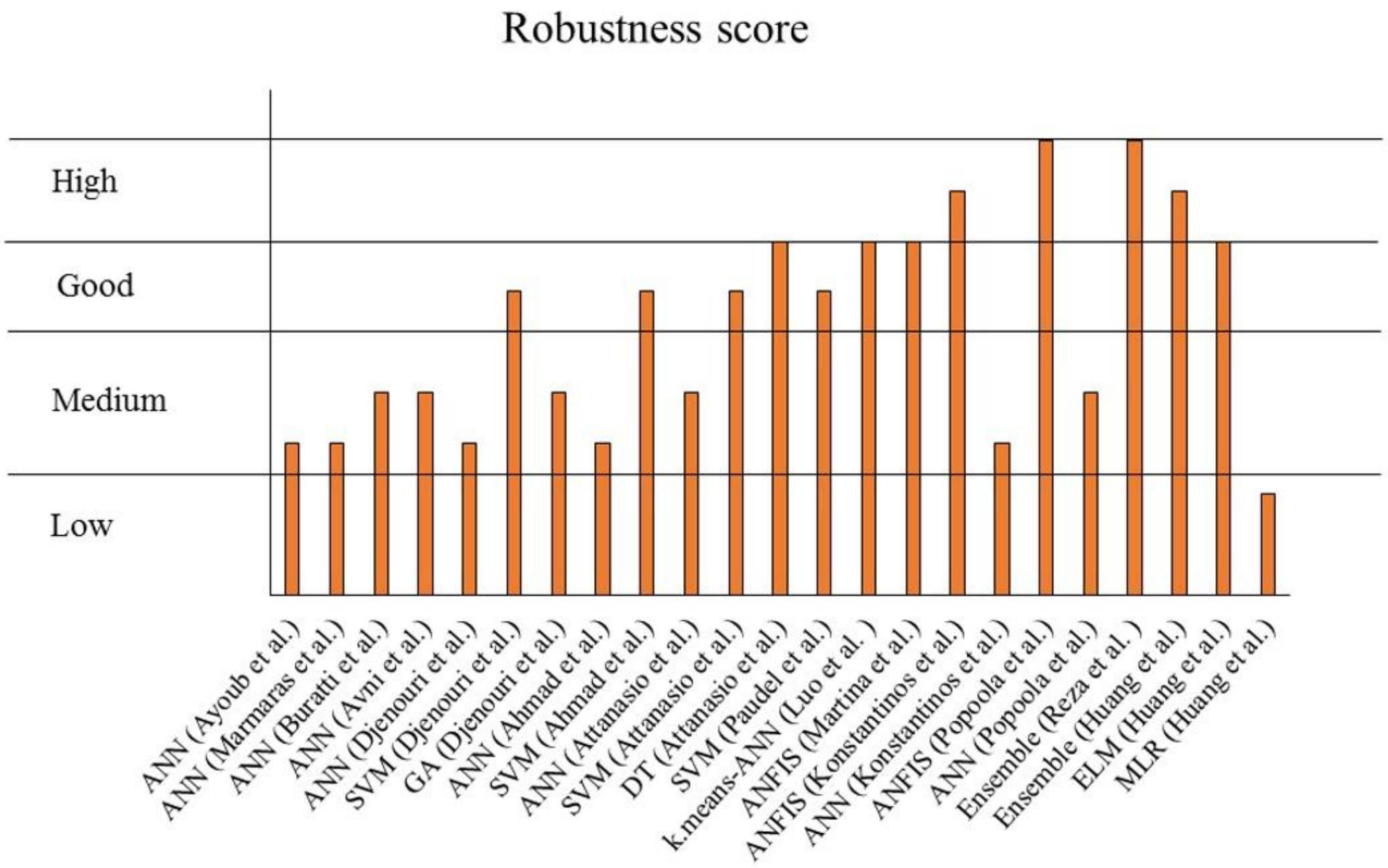

**Figure 6.** Robustness score for energy demand forecasting methods

As is clear from Fig. 6, for energy demand forecasting, hybrid and ensemble methods are located in high robustness range, SVM based methods are located in good robustness limitation, ANN based methods are located in medium robustness limitation and linear regression models are located in low robustness limitations.

## 4 BE consumption prediction

BE consumption is a factor indexed by Watt-hours which is considered as a functional unit of energy, consumed in a specific period of time (e.g., daily, monthly etc). This factor can be calculated by multiplying BE load values by the number of consumption hours (Badea, Felseghi et al. 2014). BE consumption is considered as one of the most effective factors for designing sustainable instruments in building sector and can help policy makers in taking a future perspective. Therefore, forecasting tools can play an effective role in this field. Table 4 collects and presents the most important studies developed with energy consumption prediction purposes in building sector using ML techniques.

**Table 4.** Studies developed by ML techniques for forecasting BE consumption

| Reference | Contribution | Application | ML method | Time period |
|---|---|---|---|---|
| (Finck, Li et al. 2019) | To develop a model predictive controller using ANN technique for forecasting BE consumption | Forecasting | ANN | Short term |
| (Katsanou, Alexiadis et al. 2019) | To develop ANN technique for forecasting BE consumption in term of lightening systems | Forecasting | ANN | Short term |
| (2019) (Sharif and Hammad 2019) | To present an ANN network to explore complex BE consumption in the presence of dataset extracted from the simulation-Based procedure | Forecasting | ANN | Short term |
| (Ferlito, Atrigna et al. 2015) | To develop ANN method for the prediction of BE consumption | Forecasting | ANN | Long term |
| (Chammas, Makhoul et al. 2019) | To explore for proposing an accurate ML method for the prediction of BE consumption | Forecasting | ANN, LR and SVM | Short term |
| (Zeng, Liu et al. 2019) | To develop ML based method for forecasting of electricity consumption in building | Forecasting | ANN, SVM and MLR | Short term and Long term |
| (Huang, Yang et al. 2019) | To develop SVM technique for forecasting of energy consumption for the production of hot asphalt | Forecasting | SVM | Long term |
| (Dong, Cao et al. 2005) | To develop SVM technique for the estimation of BE consumption | Forecasting | SVM | Long term |
| (Chou and Tran 2018) | To present a comprehensive review study for evaluating ML methods for the estimation of BE consumption | Forecasting | Hybrid SARIMA-MetaFA-LSSVR and SARIMA-PSO-LSSVR | Short term |
| (Solgi, Husseini et al. 2019) | To develop a hybrid hierarchical fuzzy multiple-criteria group decision making for ranking the forecasting methods of BE consumption | Clustering | hybrid hierarchical fuzzy multiple-criteria decision making | Short term |
| (Goudarzi, Anisi et al. 2019) | To develop a novel hybrid technique to estimate BE consumption from supplied data, accurately | Forecasting | Hybrid ARIMA-SVR and PSO | Short term |
| (Silva, Praça et al. 2019) | To propose ensemble methods for forecasting BE consumption | Forecasting | RF, GBRT and Adaboost | Short term |
| (Zhang, Liao et al. 2018) | To develop a study for comparing the performance of ensemble and single methods for the prediction of energy consumption in BE sector | Forecasting | RF | Long term |
| (Papadopoulos, Azar et al. 2018) | To develop tree based ensemble technique for the estimation of BE consumption | Forecasting | FR, ERT, and GBRT | Short term |

### 4.1 ANN based studies

Finck et al. (Finck, Li et al. 2019) proposed ANN method in the presence of model predictive controller for the estimation of BE consumption. ANN could successfully cope with the prediction task and could optimize the controlling process which was sustainable from economic aspect. Katsanou et al. (Katsanou, Alexiadis et al. 2019) proposed ANN technique for the estimation of the internal lighting system in a real condition in the presence of the user preferences. ANN could successfully predict the target values. In another study, Sharif and Hammad (Sharif and Hammad 2019) proposed an ANN technique for forecasting BE consumption data exported from the simulation-Based Multi-Objective

Optimization method. The study emphasizes on proposing a robust estimation technique for BE consumption. The proposed ANN technique consumes less processing time and high sustainability.

Ferlito et al. (Ferlito, Atrigna et al. 2015) developed ANN method for the prediction of the real energy demand of a generic building in the presence of monthly historical dataset related to BE consumption. Main findings have been analysed using root mean square percentage error (RMSPE). According to the findings ANN was a proper tool for forecasting energy consumption indexes.

As is clear and previously mentioned, the accuracy of ANN is not enough for covering the all dataset with different dimensions. There is a need for more accurate methods. One of these methods is support vector based technique which can be called as the frequently used techniques for the prediction of BE consumption.

### 4.2 Support vector based methods

Chammas et al. (Chammas, Makhoul et al. 2019) proposed a study for finding the estimation of the BE consumption in the presence of data exported from the IoT technique employed in building sectors. ML-based techniques containing SVM and ANN have been compared with linear regression in terms of R2, MAPE and RMSE. Database was included three main sectors including no light data, no date data and weather only data for exploring the proper variables on simulation process. According to the findings, SVM followed by ANN have higher performance compared with that for linear regression model.

Zeng et al. (Zeng, Liu et al. 2019) presented a comparative study for proposing a predictive method for forecasting electricity consumption in building sector among ANN, SVM and MLR methods. Results have been analysed using RMSE. Accordingly, SVM could successfully generate accurate results. In similar way, Huang et al. (Huang, Yang et al. 2019) proposed SVM method in comparison with kernel principal component analysis for the estimation of energy consumption for producing hot mix asphalt as a part of building sector materials. Fuel consumption and prediction error values have been employed a control values for evaluating results. SVM presented accurate results within acceptable error range. Also, Dong et al. (Dong, Cao et al. 2005) developed SVM technique for the estimation of BE consumption in China. Dataset was related to Chinese National Bureau of Statistics in thirty provinces. Evaluations for reaching a best model architecture have been performed using MSE and R2 values. Accordingly, SVM could successfully cope with the task.

### 4.3 Hybrid based methods

Chou et al. (Fenza, Gallo et al. 2019) proposed a Hybrid SARIMA-MetaFA-LSSVR and SARIMA-PSO-LSSVR techniques for the prediction of BE consumption. Results have been evaluated using correlation coefficient, RMSE, MAE and MAPE factors and compared single (ANN method), ensemble (bagging-ANN) and hybrid methods. Accordingly, the proposed hybrid methods could cope with the task as by providing the sustainability value for prediction phase. Also, it has been observed that, ensemble method has higher accuracy compared with single method and hybrid technique has higher accuracy in comparison with ensemble method.

In another study, Solgi et al. (Solgi, Husseini et al. 2019) proposed an innovative hybrid hierarchical fuzzy multiple-criteria group decision making for ranking the prediction methods in the presence of economic and environmental criteria, market related and technical advantages. The desired technique could successfully cope with the defined task. Goudarzi et al. (Goudarzi, Anisi et al. 2019) developed a novel hybrid forecasting model based on ARIMA-SVR and PSO for accurately forecasting of BE consumption in the presence of the supplied data. Results have been analysed using RMSE, MAE and MAPE. Accordingly, hybrid method could provide an accurate forecasting platform compared with PSO-SVR and ARIMA.

## 4.4 Ensemble based studies

Silva et al. (Silva, Praça et al. 2019) presented a study for the evaluation of the performance of three ensemble methods containing RF, gradient boosted regression trees and Adaboost. Evaluations propose that employing ensemble methods can be one of the proper solutions for problem rising from short term forecasting. In similar way, Zhang et al. (Zhang, Liao et al. 2018) proposed a comparison of the performance of ensemble and single techniques for proposing a proper prediction method for the prediction of the BE consumption. The developed methods were linear regression, SVM-based model, Random forest and XGBoost algorithm. Comparing have been conducted using RMSE, R2 and MAE. Accordingly, XGBoost algorithm followed by RF provided a higher performance compared with SVM and linear regression. Papadopoulos et al. (Papadopoulos, Azar et al. 2018) employed tree based ensemble methods including RF, ERTs, and GBRTs for the prediction of energy consumption in BE sector. Evaluations have been performed using MSE, MAE and MAPE factors. Accordingly, GBRTs could successfully improve the average energy consumption, significantly.

## 4.5 Results and discussions

Table 5 gives a brief comparison for the accuracy, reliability and sustainability of the models developed for forecasting the BE consumption.

**Table 5.** the comparison results of methods for BE consumption

| **Method** | **Application** | **Accuracy** | **Reliability** | **Sustainability** | **Reference** |
|---|---|---|---|---|---|
| ANN | Regression | + | + | + | (Finck, Li et al. 2019) |
| ANN | Regression | ++ | + | + | (Katsanou, Alexiadis et al. 2019) |
| ANN | Regression | + | + | + | (Sharif and Hammad 2019) |
| ANN | Regression | + | + | + | (Ferlito, Atrigna et al. 2015) |
| ANN | Regression | ++ | + | + | (Chammas, Makhoul et al. 2019) |
| LR | Regression | + | + | + | (Chammas, Makhoul et al. 2019) |
| SVM | Regression | ++ | ++ | ++ | (Chammas, Makhoul et al. 2019) |
| ANN | Regression | + | + | + | (Zeng, Liu et al. 2019) |
| SVM | Regression | ++ | ++ | ++ | (Zeng, Liu et al. 2019) |
| MLR | Regression | ++ | + | + | (Zeng, Liu et al. 2019) |
| SVM | Regression | ++ | ++ | ++ | (Huang, Yang et al. 2019) |
| SVM | Regression | ++ | ++ | + | (Dong, Cao et al. 2005) |
| SARIMA-MetaFA-LSSVR | Regression | +++ | +++ | +++ | (Chou and Tran 2018) |
| SARIMA-PSO-LSSVR | Regression | ++ | ++ | ++ | (Chou and Tran 2018) |
| HFMCD | Regression | +++ | +++ | ++ | (Solgi, Husseini et al. 2019) |
| Hybrid ARIMA-SVR-PSO | Regression | +++ | +++ | +++ | (Goudarzi, Anisi et al. 2019) |
| ARIMA | Regression | ++ | ++ | + | (Goudarzi, Anisi et al. 2019) |
| SVR-PSO | Regression | +++ | +++ | ++ | (Goudarzi, Anisi et al. 2019) |
| RF | Regression | +++ | ++ | ++ | (Silva, Praça et al. 2019) |
| GBRT | Regression | +++ | +++ | +++ | (Silva, Praça et al. 2019) |
| Adaboost | Regression | ++ | ++ | ++ | (Silva, Praça et al. 2019) |
| RF | Regression | +++ | +++ | ++ | (Zhang, Liao et al. 2018) |
| RF | Regression | +++ | ++ | ++ | (Papadopoulos, Azar et al. 2018) |
| ERT | Regression | ++ | ++ | ++ | (Papadopoulos, Azar et al. 2018) |
| GBRT | Regression | +++ | +++ | +++ | (Papadopoulos, Azar et al. 2018) |

Fig. 7 presents a concluded analytical comparison for each model according to their robustness. Fig. 7 have been separated into four limitations including high, good, medium and low robustness score to describe the capability and strength of each model according to our observations and understandings from conclusion and findings of each study.

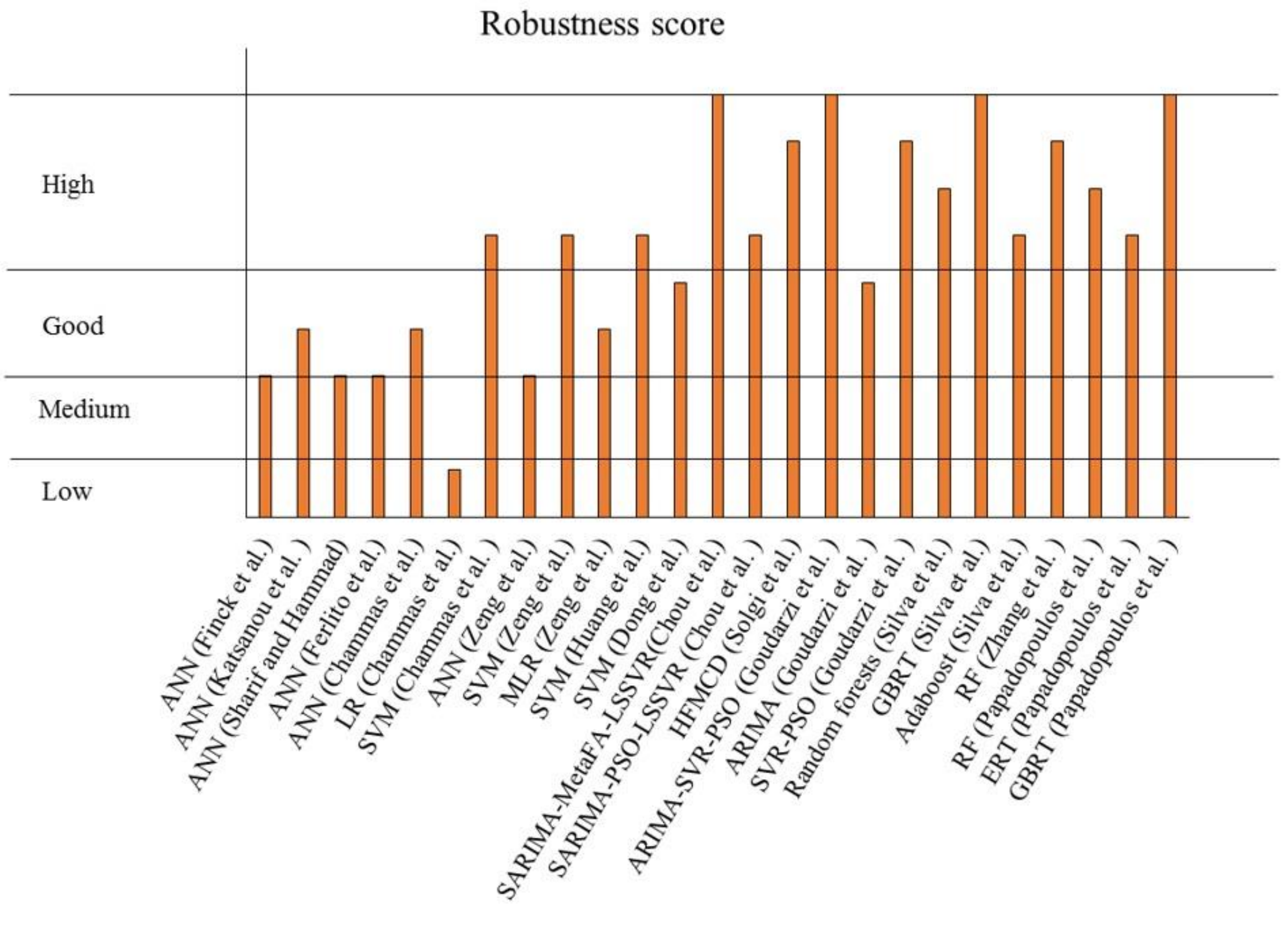

**Figure 7.** Robustness score for BE consumption prediction methods

According to Fig. 7, for energy consumption forecasting, DL-based, hybrid and ensemble based models provided the highest robustness score. ANN, SVM and single Ml-based models provided the good and medium robustness and LR-based models provided the lower robustness score.

## 5 BE load prediction

BE load is a factor indexed by Watt which is considered as a functional unit of power. This factor refers to the amount of electricity requires to operate an electrical device at any given moment (Salih 2020). BE load is another significant aspect in BE information sector. Recently, predictive methods have been employed for forecasting energy load for obtaining sustainable condition in BE sector. Table 6, presents top ML methods employed for forecasting energy load in building sector.

**Table 6.** Studies developed for forecasting energy load in building sector

| **Reference** | **Contribution** | **Application** | **ML method** | **Time period** |
| --- | --- | --- | --- | --- |
| (Dan and Phuc 2018) | To develop ANN method in comparison with MLR for the prediction of BE load | Forecasting | ANN | Short term |
| (Yuce, Mourshed et al. 2017) | To develop ANN method for the estimation of energy load in the presence of current electricity demand and social parameters | Forecasting | ANN | Long term |
| (Deb, Eang et al. 2016) | To develop ANN method for forecasting cooling load energy in building sector | Forecasting | ANN | Long term |
| (Ahmad, Chen et al. 2019) | To develop Gaussian process regression, ANN and LR to estimate the medium-term horizon cooling load in building sector | Forecasting | GPR, ANN and LR | Short term |
| (Zhu, Shen et al. 2019) | To develop ensemble ML methods for forecasting heating and cooling loads in BE sector | Forecasting | RF, SVM and LR | Long term |
| (Le, Nguyen et al. 2019) | To develop a comparative study for forecasting the heating energy load of building sectors | Forecasting | PSO-ANN, GA-ANN, ICA-ANN, and ABC-ANN | Short term |
| (Seyedzadeh, Pour Rahimian et al. 2019) | To develop Ensemble and single ML methods for the estimation of BE loads | Forecasting | SVM, RF, ANN, GBRT and XGboost | Short term |
| (Bui, Moayedi et al. 2019) | To develop a hybrid ML method for accurately prediction of heating energy load in building sector | Forecasting | ANN, M5rule, GA-M5rule | Short term |
| (Ngo 2019) | To predict different ensemble ML method for forecasting cooling energy load in building sector | Forecasting | ANN-SVR, ANN-LR, CART-SVR, CART-LR, SVR-LR, ANN-bagging and ANNs + CART + SVR | Short term |

### 5.1 ANN based studies

Dan et al. (Dan and Phuc 2018) proposed ANN method for forecasting energy load in building sector. Results of ANN method have been evaluated using RMSE factor and compared with MLR as a control method. ANN could successfully cope with forecasting task and provided a sustainable network for generating desired values. In another study, Yuce et al. (Yuce, Mourshed et al. 2017) developed ANN method for the estimation of electricity load in the presence of current energy demand and social parameters. Multiple regression analysis and principal component analysis have been employed for choosing the proper input parameters. Results have been evaluated using correlation coefficient and average percentage error values. ANN could provide acceptable output values compared with actual values. Deb et al. (Deb, Eang et al. 2016) proposed an ANN method for forecasting cooling load in building sector in the presence of energy consumption data. Results have been evaluated using R2 values. ANN could successfully cope with forecasting task by its ability to train and estimate the next day energy consumption in the presence of data related to five previous days as input variables. In another study, Ahmad et al. 2019 (Ahmad, Chen et al. 2019) developed GPR, ANN and LR techniques for forecasting the medium-term horizon cooling load in building sector in order to optimize the BE consumption. Methods have been compared in terms of correlation coefficient, MAPE and coefficient of variation. As results, GPR followed by ANN provided the highest estimation accuracy in comparison with LR.

### 5.2 Hybrid based studies

Le et al. (Le, Nguyen et al. 2019) developed hybrid ML methods for proposing an accurate predictive model for estimating the heating energy load in building sector. Results have been compared using R2, RMSE and MAE factors. Based on results, employing hybrid methods has a significant effect on increasing the accuracy of a single method. As a result, ANN-GA provided the highest performance among other techniques.

In another study, Bui et al. (Bui, Moayedi et al. 2019) developed an innovative hybrid GA-M5rule method for forecasting cooling energy load in building sector. The proposed method has been compared with ANN and M5rule in terms of R2, MAE and RMSE factors. Based on results, the proposed hybrid method could successfully increase the accuracy of forecasting.

### 5.3 Ensemble based studies

Zhu et al. (Zhu, Shen et al. 2019) developed RF method as one of the most popular ensemble methods in comparison with linear regression and SVM methods in terms of MAPE and MAE in the presence of dataset related to heating and cooling loads in building sector. Based on results, RF followed by SVM provided the highest prediction performance. Seyedzadeh et al. (Seyedzadeh, Pour Rahimian et al. 2019) developed Ensemble and single ML methods including SVM, RF, GBRT, XGboost and ANN for the estimation of BE loads in the presence of dataset related to simulated BE which have beeb generated in Energy-Plus and Ecotect. Results have been evaluated using R2, RMSE and MAE values. As a result, in complex dataset, ensemble methods provide the highest accuracy in comparison with single techniques, such that, GBRT followed by XGboost provided the highest performance and accuracies. But in simple dataset, single SVM method provided the best accuracy. Ngo (Ngo 2019) developed different ensemble methods for the estimation of cooling energy load in building sector. Results have been compared using R2, MAPE and RMSE. As results, the ensemble bagging ANN method could provide the highest prediction performance in comparison with other techniques.

### 5.4 Results and discussion

Table 7 present a brief comparison about the accuracy, reliability and sustainability of methods developed for forecasting the energy load in building sector.

**Table 7.** The comparison results of methods for energy load in building sector

| Method | Application | Accuracy | Reliability | Sustainability | Reference |
|---|---|---|---|---|---|
| ANN | Regression | + | + | + | (Dan and Phuc 2018) |
| ANN | Regression | ++ | + | + | (Yuce, Mourshed et al. 2017) |
| ANN | Regression | + | + | + | (Deb, Eang et al. 2016) |
| GPR | Regression | ++ | ++ | + | (Ahmad, Chen et al. 2019) |
| ANN | Regression | ++ | + | + | (Ahmad, Chen et al. 2019) |
| LR | Regression | + | + | + | (Ahmad, Chen et al. 2019) |
| RF | Regression | + | + | + | (Zhu, Shen et al. 2019) |
| SVM | Regression | ++ | ++ | ++ | (Zhu, Shen et al. 2019) |
| LR | Regression | + | + | + | (Zhu, Shen et al. 2019) |
| PSO-ANN | Regression | ++ | + | + | (Le, Nguyen et al. 2019) |
| GA-ANN | Regression | +++ | +++ | ++ | (Le, Nguyen et al. 2019) |
| ICA-ANN | Regression | ++ | + | + | (Le, Nguyen et al. 2019) |
| SVM | Regression | ++ | ++ | ++ | (Seyedzadeh, Pour Rahimian et al. 2019) |
| RF | Regression | +++ | +++ | ++ | (Seyedzadeh, Pour Rahimian et al. 2019) |
| ANN | Regression | ++ | + | + | (Seyedzadeh, Pour Rahimian et al. 2019) |
| GBRT | Regression | +++ | +++ | +++ | (Seyedzadeh, Pour |
| | | | | | Rahimian et al. 2019) |
| ANN | Regression | + | + | + | (Bui, Moayedi et al. 2019) |
| M5rule | Regression | ++ | ++ | ++ | (Bui, Moayedi et al. 2019) |
| GA-M5rule | Regression | +++ | +++ | +++ | (Bui, Moayedi et al. 2019) |
| ANN-SVR | Regression | ++ | ++ | ++ | (Ngo 2019) |
| ANN-bagging | Regression | +++ | +++ | +++ | (Ngo 2019) |
| ANNs + CART + SVR | Regression | +++ | +++ | ++ | (Ngo 2019) |

Fig. 8 presents a concluded indicator for each method based on their robustness. Fig. 8 have been separated into four limitations including high, good, medium and low robustness score to evaluate the capability and strength of each technique based on our own observations and understandings from conclusion and findings of each study.

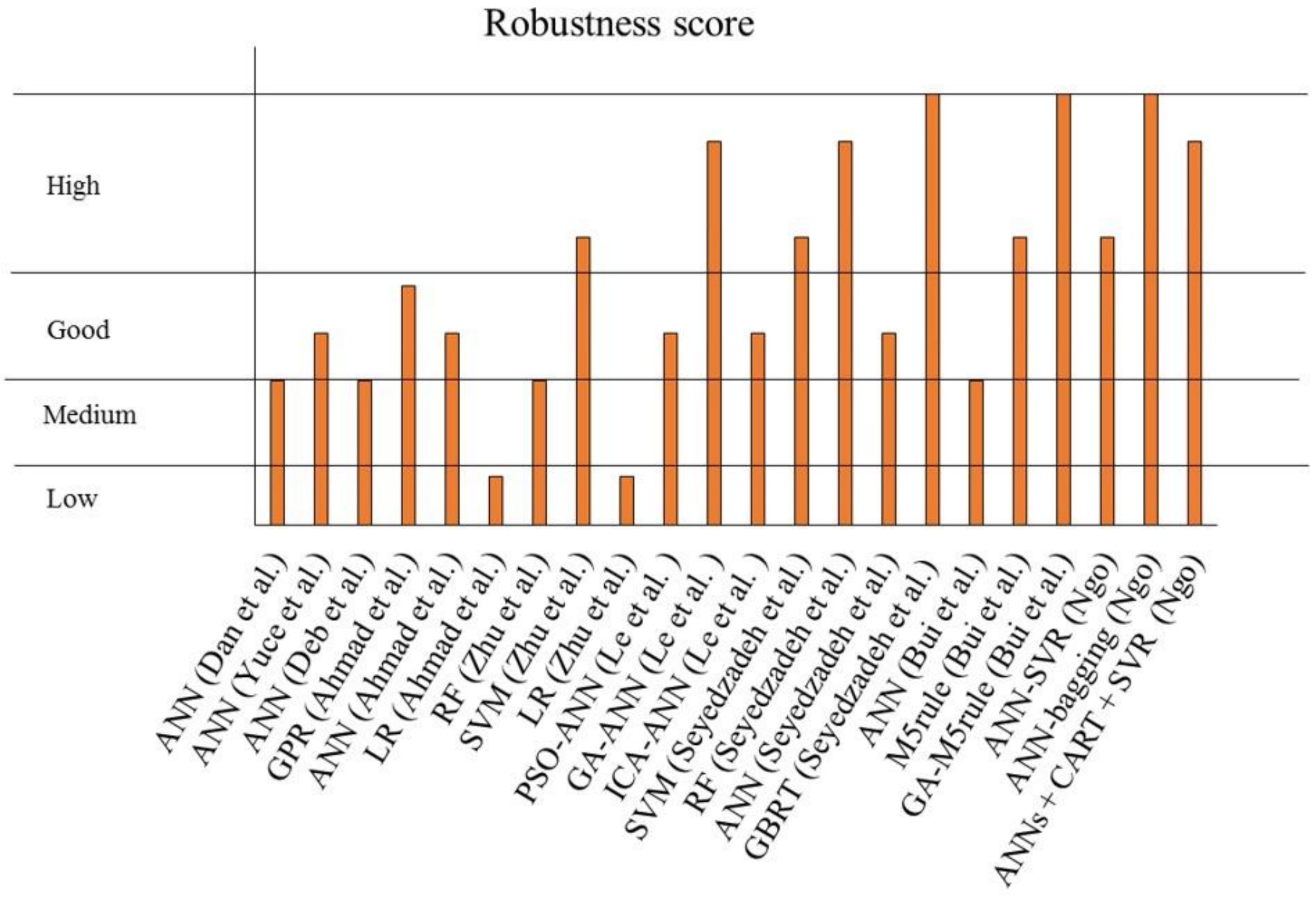

**Figure 8.** Robustness score for energy load prediction methods

According to Fig. 8, for energy load forecasting, LR based models provided the lower robustness score. The hybrid and ensemble based models provided higher robustness score. The DL-based and SVM-based techniques provided Good robustness score and ANN based techniques provided medium robustness score.

## 6 Deep learning methods in building information sector

Recently, DL techniques have been employed to be as the most accurate methods for analyzing, predicting and optimizing purposes in different fields of science. One of the most impressive fields is the building information sector. Various studies have successfully developed DL techniques for handling building information sector. Table 8 presents the most important and top researches in application of DL techniques in building information sector.

**Table 8.** Top studies for application of DL methods in building information

| Reference | Contribution | DL method | Application | Time period |
|---|---|---|---|---|
| (Wang, Hong et al. 2019) | To estimate the plug loads with occupant count data through using LSTM method in the BE sector | LSTM, ANN and LR | BE load | Short term |
| (Matsukawa, Takehara et al. 2019) | Considering the air conditioner system using LSTM method for estimation of energy consumption | LSTM | Energy consumption | Short term |
| (Singaravel, Suykens et al. 2018) | Developing LSTM method for the prediction of the energy demand in Building sector | LSTM and ANN | Energy demand | Short term |
| (Laib, Khadir et al. 2019) | to develop LSTM approach for forecasting daily gas consumption in building sector | LSTM and ANN | Energy consumption | Long term |
| (Wang, Hong et al. 2019) | To develop LSTM approach for the estimation of miscellaneous electric loads for controlling as HVAC system | LSTM | Energy load | Short term |
| (Ruiz, Capel et al. 2019) | To develop LSTM for the prediction of BE consumption | LSTM, ELMAN and NARX | Energy consumption | Long term |
| (Jain, Jain et al. 2019) | To develop a comparative study for the estimation of energy demand in building sector | LSTM | Energy demand | Long term |
| (Almalaq and Zhang 2019) | To propose an innovative hybrid LSTM-GA technique for the estimation of BE consumption | Hybrid LSTM-GA, GA-ANN and ARIMA | Energy consumption | Long term |
| (Hribar, Potočnik et al. 2019) | To estimate the residential natural gas demand in urban area using RNN | RNN and LR | Energy demand | Short term |
| (Kim, Moon et al. 2019) | Recurrent inception convolution neural network for multi short-term load forecasting | RNN, CNN and ANN | Energy load | Short term |
| (Rahman and Smith 2018) | To employ RNN technique for the estimation of energy demand in building sector | RNN and ANN | Energy demand | Short term |
| (Rahman, Srikumar et al. 2018) | To employ RNN method for the prediction of BE consumption | RNN | Electricity consumption | Long term |
| (Koschwitz, Frisch et al. 2018) | To develop RNN technique for the estimation of energy load in building sectors in comparison with SVM | SVM and NARX-RNN | Energy consumption | Short term |
| (Cai, Pipattanasomporn et al. 2019) | To propose a comparative study for the estimation of Energy load in building sector | RNN and CNN | Energy load | Long term |
| (Despotovic, Koch et al. 2019) | To develop a study for forecasting heating demand in building sector | CNN | Energy demand | Long term |
| (Zhou, Guo et al. 2019) | To propose an innovative data-driven method for the estimation of energy load in building sector | CNN | Energy load | Short term |

## 6.1 Recurrent neural networks (RNN) based studies

RNN has been developed in order to handle sequences and patterns such as text, handwriting and speech. RNN works based on cyclic connections in the structure and imports recurrent computations as input data. RNN is generally based on standard ANN that has been extended across time by having edges which feed into the next time step instead of into the next layer in the same time step. Fig. 9 shows the architecture of RNN. Additional explanations about RNN method have been presented in our previous study entitled “list of DL techniques” (Ardabili, Mosavi et al. 2019, Mosavi, Faghan et al. 2020).

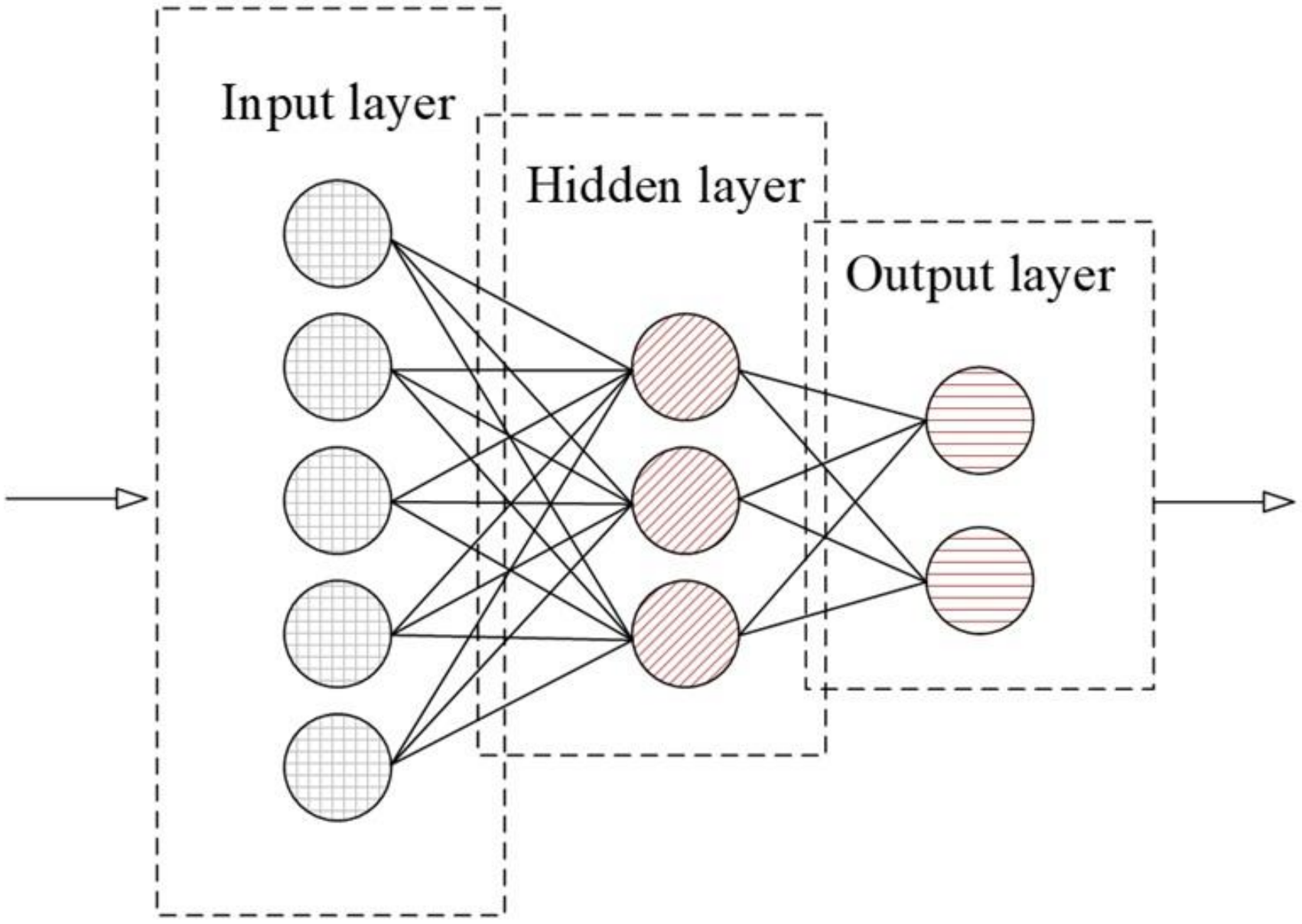

**Figure 9.** The structure of RNN method

Hribar et al. (Hribar, Potočnik et al. 2019) developed a comparative exploration for analyzing the forecasting performance of RNN and LR models for the estimation of urban natural gas demand. MAE and MAPE performance factors have been employed for evaluating the obtained results. As results RNN could provide the best performance compared with that of the LR method. Kim et al. (Kim, Moon et al. 2019) developed a RNN method for the prediction of energy load. The developed RNN method have been compared with CNN and ANN in the term of MAPE. Based on results RNN based method provided the best prediction performance.

Rahman et al. (Rahman and Smith 2018) developed RNN method in comparison with ANN technique for the prediction of energy demand in commercial buildings. Evaluations and comparison have been conducted using RMSE. RNN provided much better accuracy compared with ANN technique. In another study, Rahman et al. (Rahman, Srikumar et al. 2018) developed RNN method for the prediction of BE consumption. The developed method has been compared with ANN in terms of accuracy and robustness. As a result, RNN could improve the estimation better than ANN method.

Koschwitz et al. (Koschwitz, Frisch et al. 2018) developed RBF based SVM and Nonlinear Autoregressive Exogenous RNN for forecasting the BE load. The historical data have been employed from residential buildings in Germany in order to develop the target methods. Based on results, NARX-RNN could successfully improve the prediction performance comparing with those of SVM technique.

In another study, Cai et al. (Cai, Pipattanasomporn et al. 2019) developed a comparative study for forecasting the energy load in building sector. The developed methods were RNN and CNN. Results have been evaluated by accuracy and processing time. CNN could improve the accuracy as well as the reduction of processing time in comparison with RNN method.

### 6.2 Long short term memory (LSTM) based studies

LSTM can be considered as a subset of RNN method which is used as a general purpose computer by employing the feedback connections. The applications of LSTM are sequences and patterns recognition as well as image processing. The architecture of LSTM contains input gate, output gate and forget gate. Fig. 10 presents the architecture of LSTM method. Additional explanations about RNN method have been presented in our previous study entitled "list of DL techniques" (Ref. deep learning).

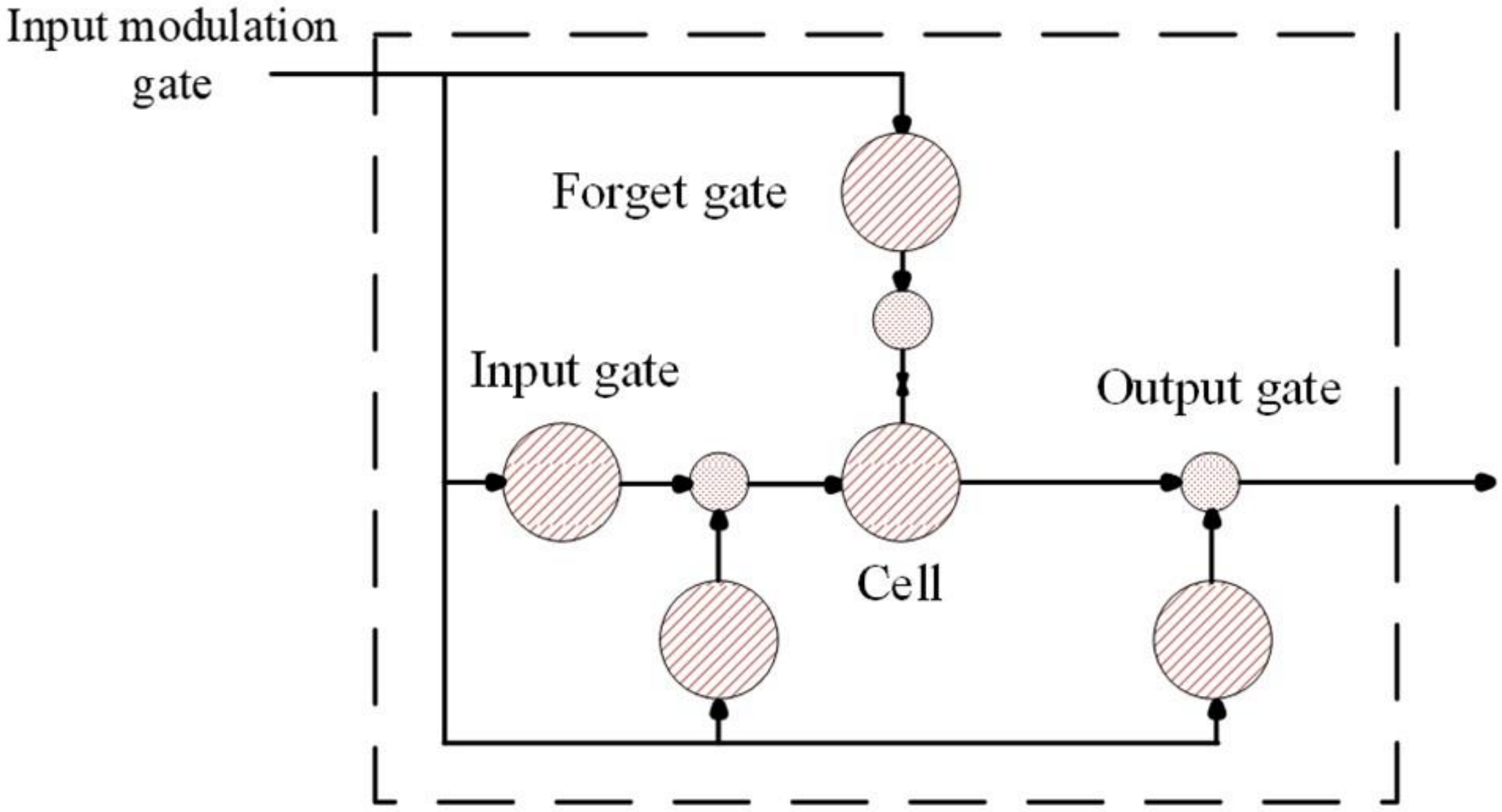

**Figure 10.** The architecture of LSTM method

Wang et al. (Wang, Hong et al. 2019) proposed LSTM technique for the estimation of BE load in comparison with LR and ANN method. Findings have been analysed using RMSE performance index. LSTM could significantly increase the prediction performance compared with LR (about 40%) and ANN (about 23.7%). Matsukawa et al. (Matsukawa, Takehara et al. 2019) explored a study for the estimation of energy consumption in air conditioner systems using LSTM methods. Using LSTM could successfully improve the prediction accuracy by up to 21%.

In another study, Singaravel et al. (Singaravel, Suykens et al. 2018) developed single, two and three layer LSTM method for forecasting BE demand factors. The developed methods have been compared with ANN in terms of accuracy and R2 values. DL techniques have been introduced to play effective roles in decreasing the processing time and enhancing the accuracy and robustness by increasing the prediction sustainability. As results, DL techniques could successfully cope with the estimation and the best prediction performance has been owned by two layered LSTM method.

Laib et al. (Laib, Khadir et al. 2019) proposed LSTM techniaue for forecasting gas consumption in building sector. Prediction results have been compared with ANN in terms of RMSE, MAE and MAPE. Results claim that, LSTM improved the prediction results about 10 to 15%.

Wang et al. (Wang, Hong et al. 2019) developed LSTM methodology for the estimation of energy load with an aim of controlling an HVAC system. Results have been analyzed by RMSE method. LSTM could successfully cope with the defined task.

Ruiz et al. (Ruiz, Capel et al. 2019) provided an exploration for the estimation of BE consumption. The developed methods were LSTM, ELMAN and NARX methods. Findings have been analyzed using RMSE values. Accordingly, LSTM could generated accurate output values and provided the highest robustness. In another study, Jain et al. (Jain, Jain et al. 2019) proposed a comparative study for the prediction of electricity demand in building sector using XGboost, ARIMA, LSTM and ANN. Evaluations and comparisons have been performed using MAPE and RMSE factors. Based on results, XGboost followed by LSTM provided the highest prediction performance compared with ARIMA and ANN. Almalaq and Zhang (Almalaq and Zhang 2019) developed a novel technique for the estimation of the BE consumption using LSTM and GA methods to obtain an evolutionary DL method. Database related to residential and commercial buildings have been employed to evaluate results. Comparison has been performed using RMSE and MAE factors. As results, the developed hybrid technique which take an evolutionary DL method improved the prediction accuracy with a high sustainability for energy consumption over the common DL methods.

### 6.3 Convolutional neural network (CNN) based studies

CNN can be considered as one of the most popular DL methods. This architecture of DL method generally handles image processing applications. CNN has three layers called convolutional, pooling and fully connected layers. In each CNN, there are two stages for training process, the feed-forward stage and the back-propagation stage. Fig. 11 presents the architecture of CNN technique. Additional explanations about RNN method have been presented in our previous study entitled “list of DL techniques” (Mosavi, Ardabili et al. 2019).

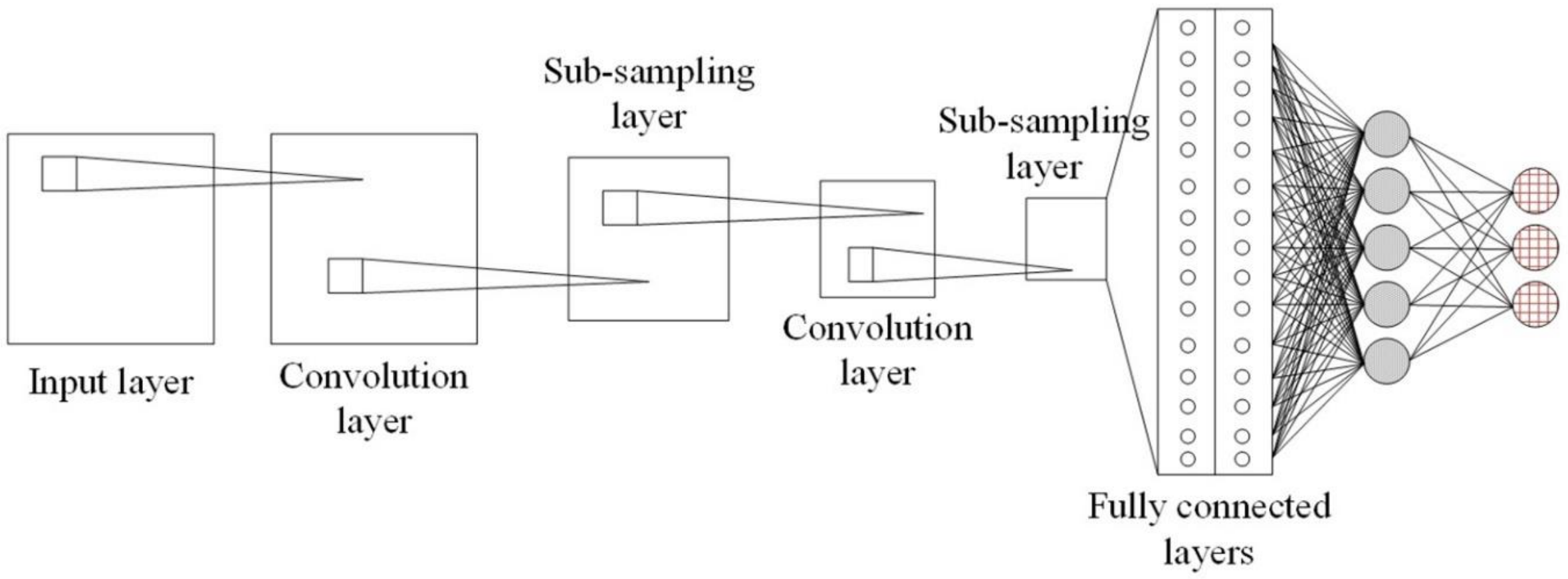

**Figure 11.** The architecture of CNN method

Despotovic et al. (Despotovic, Koch et al. 2019) developed an exploration for the estimation of heating demand in building sector using CNN technique. Findings have been conducted using accuracy and sensitivity factors. Accordingly, the developed technique could successfully estimate the heating demand with a high accuracy by 62% better than before. Zhou et al. (Zhou, Guo et al. 2019) developed a novel data-driven method using CNN technique for the prediction of energy load in building sector. Results have been evaluated using accuracy and reliability factors. As a result, CNN could successfully cope with the forecasting task with a high accuracy and sustainability.

## 6.4 Results and discussion

Table 9 gives a brief evaluation about the accuracy, reliability and sustainability of models developed for handling BE information using DL techniques.

**Table 9.** The comparison results of methods for energy information in building sector

| Method | Application | Accuracy | Reliability | Sustainability | Reference |
|---|---|---|---|---|---|
| LSTM | Regression | +++ | +++ | ++ | (Wang, Hong et al. 2019) |
| ANN | Regression | ++ | + | + | (Wang, Hong et al. 2019) |
| LR | Regression | + | - | - | (Wang, Hong et al. 2019) |
| LSTM | Regression | +++ | +++ | ++ | (Matsukawa, Takehara et al. 2019) |
| LSTM | Regression | +++ | +++ | +++ | (Singaravel, Suykens et al. 2018) |
| ANN | Regression | + | + | + | (Singaravel, Suykens et al. 2018) |
| LSTM | Regression | +++ | +++ | +++ | (Laib, Khadir et al. 2019) |
| ANN | Regression | + | + | + | (Laib, Khadir et al. 2019) |
| LSTM | Regression | +++ | +++ | ++ | (Wang, Hong et al. 2019) |
| LSTM | Regression | +++ | +++ | +++ | (Ruiz, Capel et al. 2019) |
| ELMAN | Regression | ++ | ++ | ++ | (Ruiz, Capel et al. 2019) |
| NARX | Regression | +++ | +++ | +++ | (Ruiz, Capel et al. 2019) |
| LSTM | Regression | +++ | +++ | ++ | (Jain, Jain et al. 2019) |
| LSTM-GA | Regression | +++ | +++ | +++ | (Almalaq and Zhang 2019) |
| GA-ANN | Regression | +++ | ++ | ++ | (Almalaq and Zhang 2019) |
| ARIMA | Regression | ++ | ++ | ++ | (Almalaq and Zhang 2019) |
| RNN | Regression | +++ | ++ | ++ | (Hribar, Potočnik et al. 2019) |
| LR | Regression | + | - | - | (Hribar, Potočnik et al. 2019) |
| RNN | Regression | +++ | ++ | ++ | (Kim, Moon et al. 2019) |
| CNN | Regression | +++ | +++ | +++ | (Kim, Moon et al. 2019) |
| ANN | Regression | + | + | + | (Kim, Moon et al. 2019) |
| RNN | Regression | +++ | ++ | ++ | (Rahman and Smith 2018) |
| ANN | Regression | + | + | + | (Rahman and Smith 2018) |
| RNN | Regression | +++ | +++ | ++ | (Rahman, Srikumar et al. 2018) |
| SVM | Regression | ++ | ++ | ++ | (Koschwitz, Frisch et al. 2018) |
| NARX-RNN | Regression | +++ | +++ | +++ | (Koschwitz, Frisch et al. 2018) |
| RNN | Regression | +++ | ++ | ++ | (Cai, Pipattanasomporn et al. 2019) |
| CNN | Regression | +++ | +++ | +++ | (Cai, Pipattanasomporn et al. 2019) |
| CNN | Regression | +++ | +++ | +++ | (Despotovic, Koch et al. 2019) |
| CNN | Regression | +++ | +++ | +++ | (Zhou, Guo et al. 2019) |

Fig. 12 gives a concluded evaluation for each model based on their robustness. Fig. 12 have been separated into four limitations containing high, good, medium and low robustness score to evaluate the capability and strength of each technique according to our observations and understandings from conclusion and findings of each study.

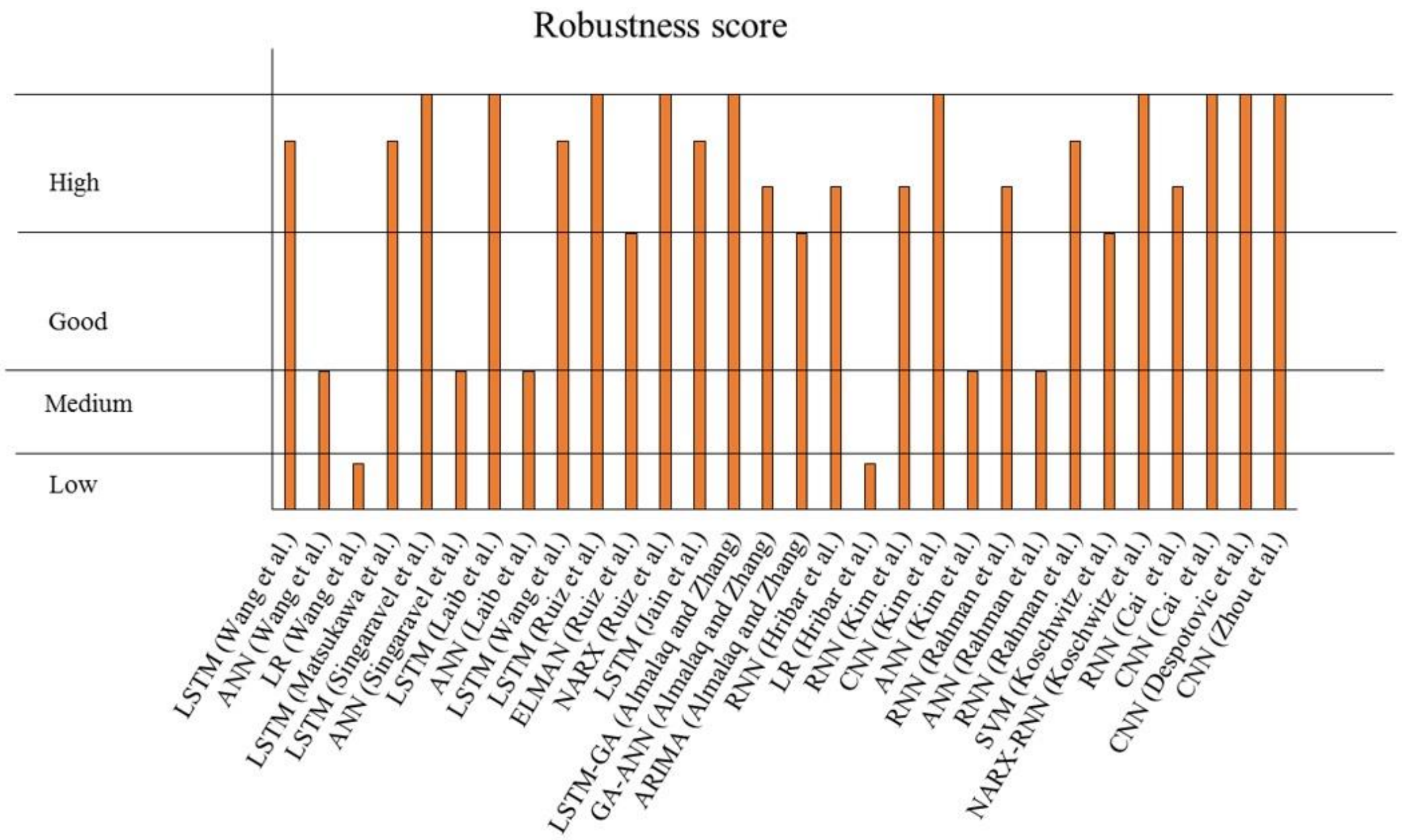

**Figure 12.** Robustness score for energy information prediction methods

According to Fig. 12, in general, DL-based techniques provided higher robustness score. On the other hand, ensemble and hybrid based models provided good robustness score but ANN and SVM based models provided medium robustness score. LR had the lowest robustness score.

## 7 Discussions

According to the results discussed in the previous section, forecasting can be considered as the major application for the use of ML-based techniques. Fig. 13 presents the trend and allocations of each application.

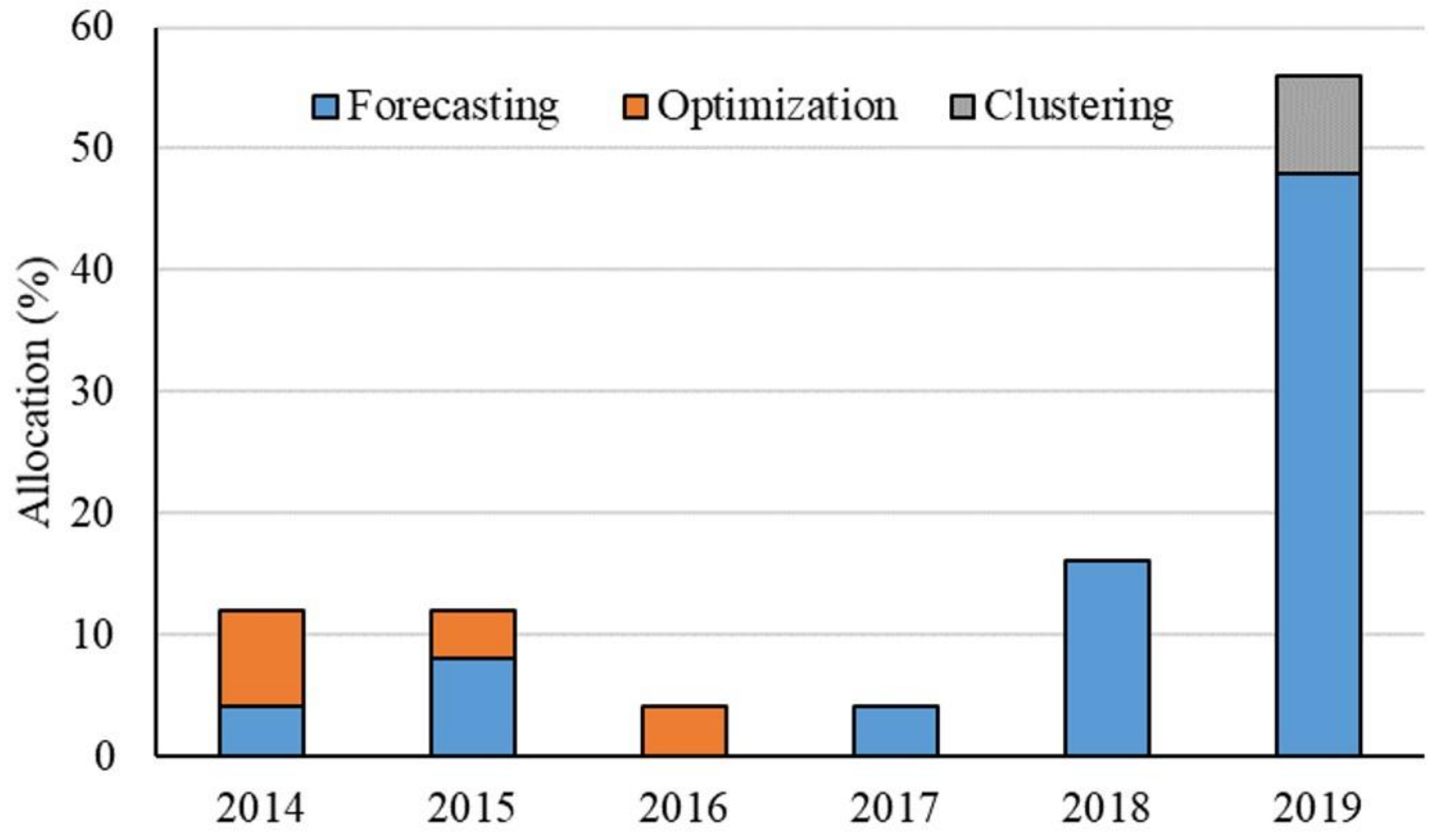

**Figure 13.** The trend and allocation of the ML-based techniques for different applications of building sector

According to Fig. 13, it can be claimed that the main trend follows Forecasting in the BE sector. As it is known, the application of AI in the fields related to the building industry is increasing, so that this increase is significant in the field of forecasting and estimation. As can be deduced from Fig. 13, most applications are related to forecasting, optimization, and clustering because these applications can be the key to entering the field of monitoring, building management, and developing intelligent security systems. On the other hand, they can also make tremendous progress in the field of smart buildings. One of the main criteria that can be discussed in this study is the issue of energy in the building. As shown in Fig. 14, the allocation of each of the models used in the topic of energy.

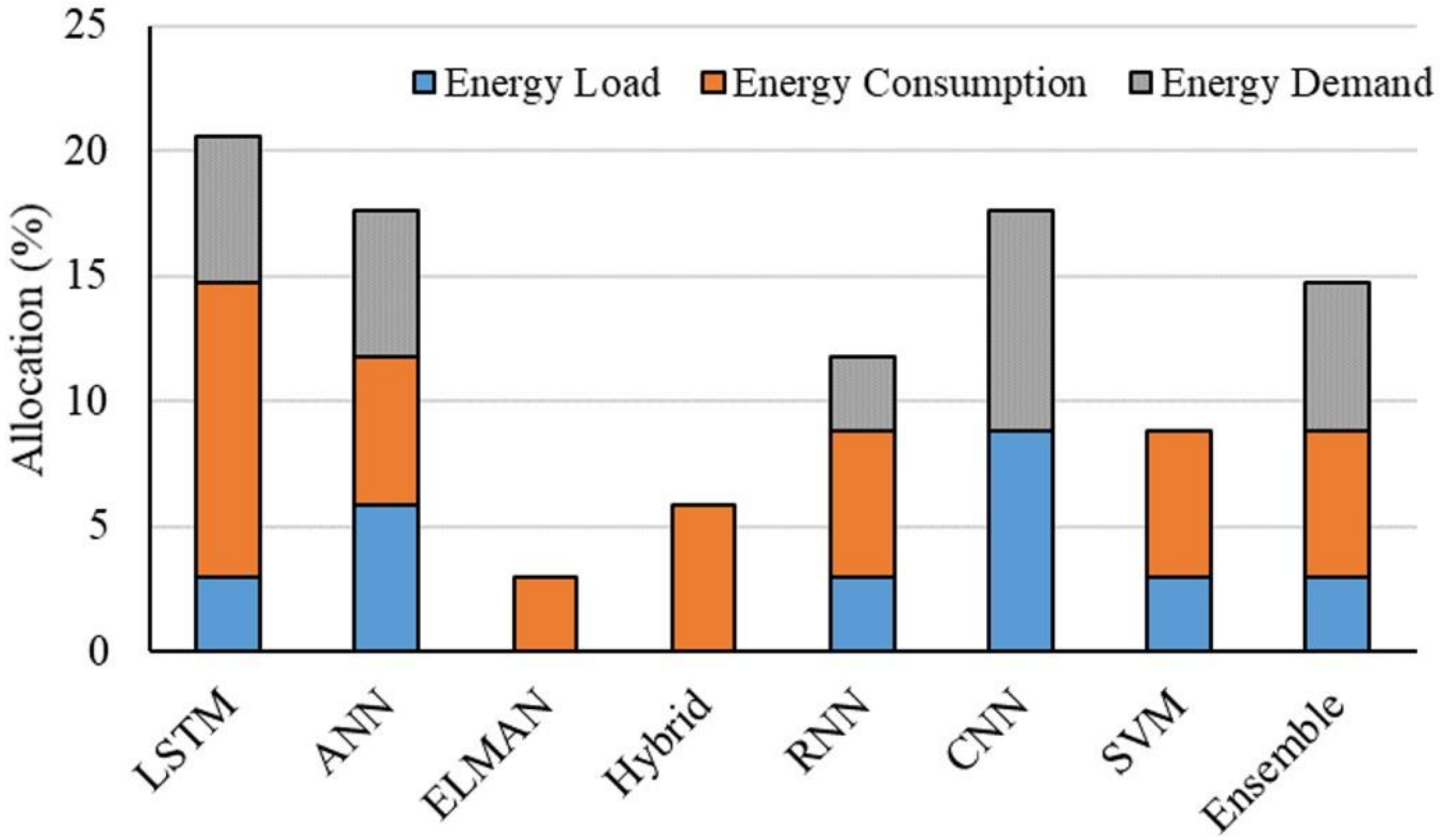

**Figure 14.** The allocation of each model in energy building

As is clear from Fig. 14, LSTM followed by ANN and CNN has the highest allocation in BE sectors. LSTM and ANN have been employed for forecasting the energy load, energy consumption, and energy demand and CNN has been employed for forecasting the energy demand and energy load. The use of each model for a specific application has some advantages and disadvantages. Table 10 presents the advantages and disadvantages of each model for forecasting applications in the BE sector.

**Table 10.** advantages and disadvantages of ML-based models in BE forecasting

| | Advantages | Disadvantages |
|---|---|---|
| LSTM | solves the problem of vanishing gradients | requires a lot of resources and time to be trained and needs high memory bandwidth |
| ANN | simple and popular and suitable for hybrid ML-based techniques | lower accuracy for huge dataset |
| ELMAN | low-dimensional data | Higher complexity |
| Hybrid | Higher accuracy compared with single model | Time consuming process |
| RNN | It can model a collection of records for estimating the patterns easily | a completely tough task to be trained |
| CNN | Can solve noises easily | lower rate of accuracy to processing time and cost |
| SVM | more effective in high dimensional spaces | It is not suitable for large data sets |
| Ensemble | ensembles provide an extra degree of freedom in the classical bias/variance tradeoff for solving problems | lack of interpretability |

According to Table 10, it can be particularly concluded that, using the DL and Ensemble techniques can be successful in case of a long-term records with an extra degree of freedom which includes a huge volume of data. But in case of small volume of data set or short-term predictions, single ML-techniques such as ANN, SVM or hybrid ML-techniques can be effective for taking a proper modeling performance in a limited process time.

The amount of energy available in a building is usually directly related to the amount of comfort of its occupants. At the same time, various categories such as environmental pollution, the volume of energy reserves at your disposal, and the costs of consumption in the form of energy. Prediction, forecasting, and estimation in the BE sector with the approach of optimization, management, and monitoring of energy availability in various applications, can be considered the point of compromise between the comfort of the building and other categories and concerns of the economic field and the environmental sustainability. The output of this study can be important from two points of view: the first is to identify weaknesses and resource management in order to move towards smart energy consumption, which can be a subset of smart buildings, and the second is to focus on smart net-zero energy or smart low-net energy buildings.

In today's world, due to the scarcity of energy resources, the importance of optimal energy consumption has become more and more important. An intelligent structure is a structure consisting of various control systems in which the activities and interactions of objects are completely intelligently managed. Therefore, a comprehensive management system and intelligent building by linking all sub-systems of the building, which unites and monitors different parts of the buildings to optimize energy consumption, improve efficiency and productivity devices, create value Increases in the building, and increases the level of comfort in the building. In this regard, the role of the Internet of Things can be more significant. Although it can be claimed that smart energy systems in the building sector are costly, the cost of this system is offset by saving energy consumption and reducing service and maintenance costs.

Smart energy systems in a building are generally defined in terms of information and communication. For example, the amount of ambient light to determine the number of lamps needed to provide lighting requires sensors to measure ambient light, a system that calculates the number of lamps required, a steering system, and a communication system to communicate between the components. The general basis for this connection in the computer world is data networks such as the Internet. Therefore, in order to achieve a large information system that includes other information such as temperature in addition to lighting data, objects work together in data networks. This interaction between objects in a computer is called the Internet of Things. Based on this, objects send and receive information with the help of predefined identifiers in the heart of Internet protocols and are in relation to the information of other objects. This information communication of objects, which is called the Internet of things, is performed through smart devices such as mobile phones. The importance of a smart energy system is also important from other aspects. In any case, energy resources are limited, and providing smart solutions for their optimal use can guarantee a better future. Depending on the texture and pattern of the city, there are numerous renewable energy sources available such as solar and wind energy, or non-renewable sources such as electricity coming from a power plant with fuel. Smart energy systems can be considered in the form of confronting the mentioned limitations in order to increase the level of satisfaction and at the same time reduce the cost. For example, the existence of smart curtains to deal with the sun and its effect on reducing the temperature of the building in summer or vice versa, using it to deal with winter cold along with other factors such as the amount of light required by the building can be an example of the category Be. This level of intelligence, in addition to preventing energy waste, can increase the level of satisfaction as well as increase the life of construction equipment, reduce the time required for service, and reduce energy costs. And therefore considered part of the principles of building engineering. It is necessary to explain that intelligence is not necessarily based on electronic and computer equipment, and the implementation of intelligent plans in building construction or intelligent selection of building equipment and facilities can also be discussed and accepted as intelligence. Future studies are moving towards the use of the Internet of Things in the smart energy systems in the building sector.

## 8 Conclusion

This study explores the usage of ML-based techniques in BE information applications. The ensemble and hybrid techniques have emerged and continue to advance for higher accuracy and better performance. DL-based techniques also will bring tremendous amount of intelligence for improving the prediction techniques. The following findings can be concluded.

- For energy demand forecasting, hybrid and ensemble methods are located in high robustness range, SVM based methods are located in good robustness limitation, ANN based methods are located in medium robustness limitation and linear regression models are located in low robustness limitations.

- For energy consumption forecasting, DL-based, hybrid and ensemble based models provided the highest robustness score. ANN, SVM and single Ml-based models provided the good and medium robustness and LR-based models provided the lower robustness score.

- For energy load forecasting, LR based models provided the lower robustness score. The hybrid and ensemble based models provided higher robustness score. The DL-based and SVM-based techniques provided Good robustness score and ANN based techniques provided medium robustness score.

- In general, DL-based techniques provided higher robustness score. On the other hand, ensemble and hybrid based models provided good robustness score but ANN and SVM based models provided medium robustness score. LR had the lowest robustness score.

The importance of a smart energy system is also important from other aspects. In any case, energy resources are limited, and providing smart solutions for their optimal use can guarantee a better future. Depending on the texture and pattern of the city, there are numerous renewable energy sources available such as solar and wind energy, or non-renewable sources such as electricity coming from a power plant with fuel. Future studies are moving towards the use of the Internet of Things in the smart energy systems in the building sector. On the other hand, due to advances in ML-based techniques, there is still a computational cost due to the high complexity of data in building optimization and energy management problems, including multistage and nonlinear behavior of building thermal performance, discontinuity in optimization variables. Uncertainty in building processes and design parameters is a major barrier to the widespread use of soft computing. Future studies should focus on developing new methods and efficient solutions based on DL and ensemble methods and using them realistically on optimization studies and evaluating their performance and tools. The energy management sector in the building, according to the volume of data obtained and also its impact on energy consumption, energy load and energy demand, as the core of energy, is the main goal of future studies.

## 9 Nomenclature

| | | | |
|---|---|---|---|
| Artificial neural network | ANN | Feed-forward neural networks | FFNN |
| Adaptive neuro fuzzy inference system | ANFIS | Particle swarm optimization | PSO |
| Analytic network process | ANP | Random forest | RF |
| Autoregressive integrated moving average | ARIMA | Centroid mean | CM |
| Building Energy | BE | Deep learning | DL |
| Extreme learning machine | ELM | Non-random two-liquid | NRTL |
| Decision tree | DT | Recurrent neural network | RNN |
| Support vector machine | SVM | Partial least squares | PLS |
| Wavelet neural networks | WNN | Discriminant analysis | DA |
| Support vector regression | SVR | Principal component analysis | PCA |
| Genetic algorithm | GA | Linear discriminant analysis | LDA |
| Multi layered perceptron | MLP | Radial basis function | RBF |
| Long short-term memory | LSTM | Extremely randomized trees | ERT |
| Machine learning | ML | Mean absolute percentage error | MAPE |
| Response surface methodology | RSM | Multi criteria decision making | MCDM |
| Back propagation neural network | BPNN | Genetic programming | GP |
| Least-squares | LS | Multi linear regression | MLR |
| Sparse Bayesian | SB | Step-wise Weight Assessment Ratio Analysis | SWARA |
| Generalized boosted regression | GBR | Multi Objective Optimization by Ratio Analysis | MOORA |
| Gaussian process regression | GPR | Nonlinear autoregressive exogenous | NARX |
| Recurrent neural networks | RNN | hierarchical fuzzy multiple-criteria decision making | HFMCD |
| gradient boosted regression trees | GBRT | seasonal autoregressive integrated moving average | SARIMA |
| Artificial bee colony | ABC | Determination coefficient | R2 |
| Classification and regression tree | CART | Mean absolute error | MAE |
| Convolutional neural network | CNN | Root mean square error | RMSE |

## Conflict of Interest

The authors declare that the research was conducted in the absence of any commercial or financial relationships that could be construed as a potential conflict of interest.

## Author Contributions

Conceptualization, S.A and A. M.; Methodology, S.A. and A.M.; Investigation, L.A.; software, S.A.; formal analysis, S.A. and A.M.; writing—original draft preparation, S.A.; visualization, S.A, L.A. and A.M.; supervision, S.A., C.M., B.T., and A.M. All authors have read and agreed to the published version of the manuscript.

## ACKNOWLEDGMENT

This project has partly been supported by funding from the European Union´s Horizon 2020 Research and Innovation Programme under the Programme SASPRO 2 COFUND Marie Sklodowska-Curie grant agreement No. 945478.